\documentclass[journal]{IEEEtran}
\usepackage{amsmath,graphicx}
\usepackage{amssymb}
\usepackage[printonlyused,nolist]{acronym}
\usepackage{pgfplots}
\usepackage{cite}
\usepackage{pgfplotstable}
\usepackage{psfrag} 
\usepackage{tikz}
\usepackage{textcomp}
\usepackage{array,multirow}
\usepackage{booktabs}
\usepackage{flushend} 
\definecolor{dunkelgruen}{rgb}{0,0.6,0} 
\definecolor{dunkelblau}{rgb}{0,0,1}

\tikzset{dashdot/.style={dash pattern=on .8pt off 3pt on 4pt off 3pt}}
\newcommand{\specialcell}[2][c]{\begin{tabular}[#1]{@{}c@{}}#2\end{tabular}}
  \usepackage{subfigure}

\begin{document}

\begin{acronym}
\acro{AEC}{acoustic echo cancellation}
\acro{NLMS}{normalized least mean squares}
\acro{PDF}{probability density function}
\acro{NLMS}{normalized least mean square}
\acro{SA}{significance-aware}
\acro{RIR}{room impulse response}
\acro{EPFES}{elitist particle filter based on evolutionary strategies}
\acro{FIR}{finite impulse response}
\acro{MMSE}{minimum mean square error}
\acro{UNGM}{Univariate Nonstationary Growth Model}

\end{acronym}
%
\title{\huge Estimating parameters of nonlinear systems using~the elitist particle filter based on~evolutionary strategies}
\author{Christian Huemmer,
        Christian~Hofmann,
	Roland Maas,
        and~Walter~Kellermann
\thanks{\noindent Multimedia Communications
and Signal Processing, Friedrich-Alexander University Erlangen-N\"urnberg (FAU), Cauerstr. 7, 91058 Erlangen, Germany, e-mail: \{huemmer,hofmann,maas,wk\}@LNT.de, R. Maas was with the FAU while the work has been conducted. He is now with Amazon, Seattle, WA.}
\thanks{The authors would like to thank the Deutsche Forschungsgemeinschaft (DFG) for supporting this work (contract number KE 890/4-2).}}

\markboth{May~2016}%
{Shell \MakeLowercase{\textit{et al.}}: Bare Demo of IEEEtran.cls for Journals}
\maketitle

\begin{abstract}
In this article, we present the elitist particle filter based on evolutionary strategies (EPFES) as an efficient approach for nonlinear system identification.
The EPFES is derived from the frequently-employed state-space model, where the relevant information of the nonlinear system is captured by an unknown state vector.
Similar to classical particle filtering, the EPFES consists of a set of particles and respective weights which represent different realizations of the latent state vector and their likelihood of being the solution of the optimization problem.
As main innovation, the EPFES includes an evolutionary elitist-particle selection which combines long-term information with instantaneous sampling from an approximated continuous posterior distribution.
In this article, we propose two advancements of the previously-published elitist-particle selection process. Further, the EPFES is shown to be a generalization of the widely-used Gaussian particle filter and thus evaluated with respect to the latter for two completely different scenarios:
First, we consider the so-called univariate nonstationary growth model with time-variant latent state variable, where the evolutionary selection of elitist particles is evaluated for non-recursively calculated particle weights.
Second, the problem of nonlinear acoustic echo cancellation is addressed in a simulated scenario with speech as input signal:
By using long-term fitness measures, we highlight the efficacy of the well-generalizing EPFES in estimating the nonlinear system even for large search spaces.
Finally, we illustrate similarities between the EPFES and evolutionary algorithms to outline future improvements by fusing the achievements of both fields of research.
\end{abstract}


\IEEEpeerreviewmaketitle

\section{Introduction}
\IEEEPARstart{T}{he} identification of time-varying nonlinear systems based on noisy measurements has been investigated for several decades and is still an active research topic~\cite{novak2014,markonmano2014,ChengPu2014}. The main challenge of such supervised identification tasks are unknown systems of nonlinear characteristics which preclude analytical solutions of statistically optimal estimation techniques, e.g. based on the \ac{MMSE} criterion.
This is why a variety of different approaches have been proposed, which can be categorized in the following way:
First, the nonlinear characteristics of the unknown system are deterministically approximated using local linearisation or functional approximation techniques to derive analytically tractable estimation algorithms. Very popular examples for these kinds of algorithms are
the extended Kalman filter \cite{daum2005}, Kernel methods~\cite{KernelMachine,KernelLearning,KernelBook} and estimation techniques for Hammerstein group models \cite{KuechPower,malik2011,Hofmann} and Volterra \mbox{filters~\cite{VFilter1,VFilter2,VFilter3}}.
In the second category, an analytically-intractable estimation technique is approximated using sequential Monte Carlo approaches, like particle filters~\cite{doucet_s2000,MCMC,cappe2007,fearnhead2008}. These numerical sampling methods capture nonlinear relations between non-Gaussian-distributed random variables by representing a probability distribution with a finite set of particles and respective weights (likelihoods).
On the one hand, such particle filters allow for a flexible and mathematically complex way of modeling the nonlinear dynamics of an unknown system. On the other hand,
particle filters are computationally expensive as a high estimation accuracy requires a large number of particles. However, many real-time implementations of particle filtering in the fields of tracking, robotics or biological engineering highlight the increased applicability with growing computational power and the use of parallel processing units~\cite{pupilli2005,gokdemir2009,Atia2010,lee2010,henriksen2012,huang2013,chitchian2013}.\\
The classical particle filter (proposed more than 20 years ago~\cite{Particle2}) has been refined in a variety of publications, see overview articles such as \cite{crisan2002,Particle2002,andrieu2004,doucet_t2009}. 
In general, particle filtering is derived from the well-known state-space model, where the relevant information about the unknown system is modeled by a latent state vector and where the posterior distribution of the state vector is modeled as a discrete \ac{PDF} (consisting of a finite set of particles and respective weights). This leads to the approximation of statistically optimal estimation techniques and the well-known problems of degeneracy and sample impoverishment, where the set of particles is represented by a few members with large weights and many particles with negligible weights.
To address these issues, different concepts have been proposed which can be assigned to two categories, namely resampling methods~\cite{Res2004,bolic2005,Xiaoyan2010} and posterior-distribution approximations (by using a continuous \ac{PDF}, see~\cite{GPF}).\\
In this article, we present and advance the \ac{EPFES} which selects elitist particles by means of long-term fitness measures and introduces innovation into the set of particles by sampling from an approximated continuous posterior \ac{PDF}~\cite{epfes}.
As the \ac{EPFES} will be shown to be a generalization of the widely-used Gaussian particle filter~\cite{GPF} (this relation has not been considered so far), we evaluate the conceptional differences with respect to the latter for two completely different state-space models. Thereby, the \ac{EPFES} is shown to
achieve a remarkable performance for the practically-relevant case of a small number of samples and to outperform the Gaussian particle filter for two completely different scenarios.
Finally, we consider the \ac{EPFES} from a different point of view and show similarities to basic features of evolutionary algorithms. This comparison motivates further improvements of the \ac{EPFES} by fusing the advantages of both concepts. \pagebreak \newline
This article is organized as follows:
In Section~\ref{sec:SSmodel}, we briefly review classical particle filtering from a Bayesian network perspective and emphasize how it differs from linear adaptive filtering concepts like the \ac{NLMS} algorithm.
This is followed by the derivation and advancement of the \ac{EPFES} in Section~\ref{cha:EPFES}, where we introduce the \ac{EPFES} as generalization of Gaussian particle filtering. 
The experimental evaluation in Section~\ref{cha:Experiments} is divided into two parts: first, we consider the \ac{UNGM}
to study the tracking capabilities of the \ac{EPFES} without using long-term information. Second, we address the task of nonlinear \ac{AEC}, where we identify nonlinear loudspeaker signal distortions by exploiting the
generalization properties of the \ac{EPFES}, improving the system identification performance even for large search spaces.
This is followed by an outlook in Section~\ref{cha:outlook}, where we motivate further refinements of the proposed elitist-particle selection of the \ac{EPFES} by highlighting parallels to evolutionary algorithms. Finally, conclusions are drawn in Section~\ref{cha:concl}.
\section{Classical particle filtering for nonlinear system identification}
\label{sec:SSmodel}
\begin{figure}[tb]
\centering
\includegraphics[width=0.35\textwidth]{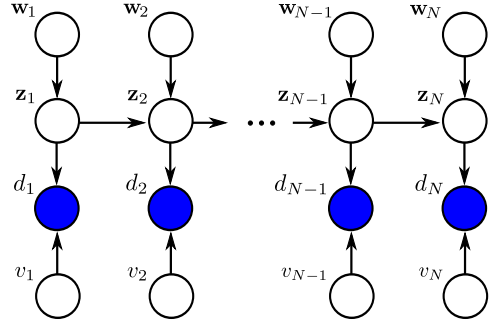}
\caption{Bayesian network of a state-space model}
\label{fig:ParticleModel}
\end{figure}

\begin{figure}[t]
\input{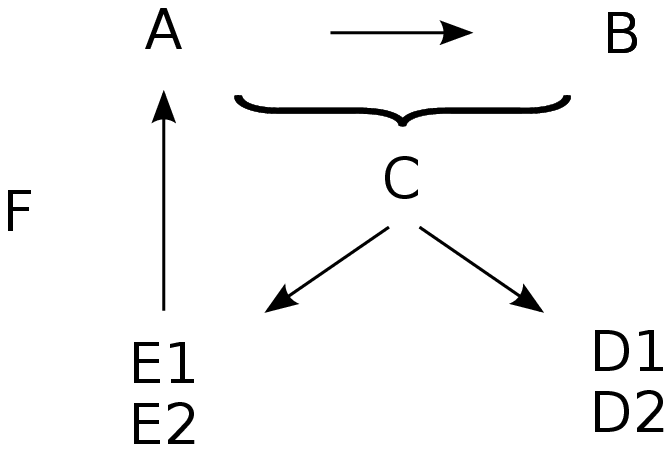}	
\caption{Concept of the classical particle filter}
\vspace{-2mm}
\label{fig:ParticleAlg}
\end{figure}

Throughout this article, the Gaussian probability density function (PDF) of a real-valued length-$M$ random
vector with mean vector $\boldsymbol{\mu}_{\mathbf{z},n}$
and covariance matrix $\mathbf{C}_{\mathbf{z},n}$, dependent on the time instant $n$,
is denoted as
\begin{equation}
\begin{split}
\mathbf{z}_n& \sim \mathcal{N}\{\boldsymbol{\mu}_{\mathbf{z},n},\mathbf{C}_{\mathbf{z},n}\}\\[1mm]
=&\frac{|\mathbf{C}_{\mathbf{z},n}|^{-1/2}}{(2\pi)^{M/2}} \exp\left\{ -\frac{(\mathbf{z}_n-\boldsymbol{\mu}_{\mathbf{z},n})^\text{T} \mathbf{C}_{\mathbf{z},n}^{-1} (\mathbf{z}_n-\boldsymbol{\mu}_{\mathbf{z},n})}{2}\right\},
\end{split}
\end{equation}
where $|\cdot|$ represents the determinant of a matrix. Furthermore, $\mathbf{C}_{\mathbf{z},n}=C_{\mathbf{z},n} \mathbf{I}$
(with identity matrix $\mathbf{I}$)
implies the elements of $\mathbf{z}_n$ to be mutually statistically independent and of equal variance~$C_{\mathbf{z},n}$.\\
For clarity, we make the following additional assumptions with respect to the experimental verification in Section~\ref{cha:Experiments}: First, we restrict the uncertainties in the process and the observation equation to be additive random variables, and the observation $d_n$ to be a scalar real-valued random variable.
Second, all vectors have real-valued entries.\\
In the following, assume the relevant information of a nonlinear system to be captured by the length-$R$ state vector
\begin{equation}
\mathbf{z}_n=[z_{0,n},z_{1,n},...,z_{R-1,n}]^\text{T}
\label{equ:zDef}
\end{equation}
with coefficients $z_{\nu,n}$ and ${\nu=0,...,R-1}$. This unobservable or latent vector depends on the time instant ${n=1,...,N}$ and its 
temporal evolution is described by the process equation
\begin{equation}
\mathbf{z}_n = \mathbf{f} \left(\mathbf{z}_{n-1} \right) + \mathbf{w}_n,
\label{equ:TranEq}
\end{equation}
where $\mathbf{f}\left( \cdot \right)$ represents the so-called nonlinear progress \cite{Particle2}.
The uncertainty of the state vector is denoted as $\mathbf{w}_n$ and is of same dimension
as $\mathbf{z}_n$ in (\ref{equ:TranEq}).
The relationship between the state vector $\mathbf{z}_n$ and the observation $d_n$ is described by
\begin{equation}
d_n = \mathbf{g} \left( \mathbf{x}_n,\mathbf{z}_n \right) +v_n,
\label{equ:ObservGen}
\end{equation}
where $\mathbf{g}(\cdot)$ represents a nonlinear function which also depends on the length-$M$ input signal vector
\begin{equation}
\mathbf{x}_n=[x_n,x_{n-1},...,x_{n-M+1}]^\text{T}
\end{equation}
with time-domain samples $x_n$.
Furthermore, the uncertainty of the observation $d_n$ in (\ref{equ:ObservGen}) is modeled by the additive variable~$v_n$.
Throughout this article, we assume $\mathbf{w}_n$ and $v_n$ to be normally distributed and of zero mean:
\begin{equation}
\mathbf{w}_n \sim \mathcal{N}\{ \mathbf{0},\mathbf{C}_{\mathbf{w},n}\}, \quad v_n \sim \mathcal{N}\{ 0,\sigma^2_{v,n}\}.
\label{equ:Gaussv}  
\end{equation}
From a Bayesian network perspective, this probabilistic model corresponds to the graphical model shown in Fig.~\ref{fig:ParticleModel}, where
the observed input signal vector $\mathbf{x}_n$ is omitted for notational convenience in the later probabilistic calculus ($\mathbf{x}_n$ specifies the value of the function $\mathbf{g}(\cdot)$ and is thus captured by $d_n$).
In Fig.~\ref{fig:ParticleModel}, the directed links express statistical dependencies between the nodes whereas observed variables, such as $d_n$, are marked by shaded circles.
The estimate of the state vector $\mathbf{\hat{z}}_n$
is derived~as
\begin{equation}
\mathbf{\hat{z}}_n = \underset{\mathbf{\tilde{z}}_n}{\operatorname{argmin}} \; \mathcal{E} \{ ||\mathbf{\tilde{z}}_n-\mathbf{z}_n ||_2^2 \},
\label{equ:mes}
\end{equation}
reflecting the \ac{MMSE} criterion, where $||\cdot||_2$ is the Euclidean norm and $\mathcal{E}\{\cdot\}$ the expectation operator. As characteristic for Bayesian~\ac{MMSE} estimation, the minimization of (\ref{equ:mes}) with respect to $\mathbf{\tilde{z}}_n$
yields the mean vector of the posterior \ac{PDF} $p \left( \mathbf{z}_n | d_{1:n}\right)$ as estimate for the state vector
\begin{equation}
\mathbf{\hat{z}}_n= \mathcal{E} \{ \mathbf{z}_n | d_{1:n} \},
\label{equ:zHatE}
\end{equation}
where ${d_{1:n}=d_1,...,d_n}$.
In the case of linear relations between the variables in (\ref{equ:TranEq})~and~(\ref{equ:ObservGen}),
and for a linear estimator for $\mathbf{\hat{z}}_n$, the \ac{MMSE}~estimate of (\ref{equ:mes}) leads to the Kalman filter equations~\cite{Bishop} and - under further assumptions on the statistics of the random variables - to the NLMS algorithm with an adaptive stepsize value~\cite{Huemmer2014}.\\
However, if the process equation~(\ref{equ:TranEq})~or the observation equation~(\ref{equ:ObservGen}) exhibit nonlinear structure,
we cannot derive the Bayesian estimate of $\mathbf{\hat{z}}_n$ in~(\ref{equ:zHatE}) in a closed-form analytical way.
Thus, we employ the particle filter to approximate the posterior \ac{PDF}\\
\begin{equation}
p \left( \mathbf{z}_n | d_{1:n}\right) = \frac{p (d_{n}|\mathbf{z}_n ) p ( \mathbf{z}_n | d_{1:n-1}) }{ \int{p (d_{n}|\mathbf{z}_n ) p ( \mathbf{z}_n | d_{1:n-1}) \text{d}\mathbf{z}_n}  }  
\end{equation}
by a discrete distribution~\cite{Bishop,PFevol}
\begin{align}
p ( \mathbf{z}_n | d_{1:n})  &\approx \sum\limits_{l=1}^L  \frac{p( d_n|  \mathbf{z}^{(l)}_n ) \delta ( \mathbf{z}_n - \mathbf{z}^{(l)}_n )}{\sum\limits_{l=1}^L \int p( d_n|  \mathbf{z}^{(l)}_n ) \delta ( \mathbf{z}_n-\mathbf{z}^{(l)}_n )\text{d}\mathbf{z}_n} \\
&=\sum\limits_{l=1}^L \omega^{(l)}_n  \delta ( \mathbf{z}_n - \mathbf{z}^{(l)}_n ),
\label{equ:DiscretePDF}
\end{align}
where $\delta \left(\cdot \right)$ is the Dirac delta distribution and $l=1,...,L$.
Based on (\ref{equ:DiscretePDF}), the set of $L$ particles $\mathbf{z}^{(l)}_n$ is characterized by the weights
\begin{equation}
\omega^{(l)}_n  = \frac{p( d_n|  \mathbf{z}^{(l)}_n  )}{\sum\limits_{l=1}^L p( d_n| \mathbf{z}^{(l)}_n  )},
\label{equ:weights}
\end{equation}
which describe the likelihoods that the observation is obtained by the corresponding particle. These likelihoods are used as measures for the probability of the samples 
to be drawn from the true \ac{PDF} \cite{Schoen}. Finally, the estimate for the state vector
\begin{equation}
\mathbf{\hat{z}}_n= \mathcal{E} \{ \mathbf{z}_n | d_{1:n} \} \approx \sum_{l=1}^L \omega^{(l)}_n   \mathbf{z}^{(l)}_n
\label{equ:ParticleFilUpdate1}
\end{equation}
is given as the estimated mean vector of the approximated posterior \ac{PDF}.
This fundamental concept is illustrated in Fig.~\ref{fig:ParticleAlg} and follows the explanation in~\cite{epfes}.
As starting point of this iterative estimation method, $L$ samples $\mathbf{z}^{(l)}_0$ are drawn from the initial~\ac{PDF} $p\left( \mathbf{z}_0 \right)$ and inserted into the time update of~(\ref{equ:TranEq}) to determine the particles $\mathbf{z}^{(l)}_1$.
Then, we calculate the particle weights $\omega^{(l)}_1$ following~(\ref{equ:weights}). With this set of particles and their associated weights, the estimate $\mathbf{\hat{z}}_1$
and the discrete distribution $p \left( \mathbf{z}_1 | d_{1}\right)$ are determined based on
(\ref{equ:ParticleFilUpdate1})~and~(\ref{equ:DiscretePDF}), respectively.
This \ac{PDF} is the basis for the sampling procedure of the following time step, in which the estimation procedure is repeated.
\noindent As shown by the circular structure in Fig.~\ref{fig:ParticleAlg}, the starting point for classical particle filtering is a
finite number of $L$ samples $\mathbf{z}^{(l)}_n$ which are subsequently evaluated in terms of their likelihood producing the observations $d_n$~\cite{Particle2002}.
This leads to three problems (Pr1, Pr2 and Pr3) which will be summarized in the following.\\[1mm]
\hspace*{3mm}\textit{Pr1 - High computational load:}
It is well known in the particle-filtering community that both estimation accuracy and computational cost increase with the number of $L$ particles~\cite{Particle2002,andrieu2004}. One approach to avoid a dissatisfying tradeoff is the selection of elitist particles by dropping all particles with weights smaller than a specific threshold~\cite{cho_2014}.\\[1mm]
\hspace*{3mm}\textit{Pr2 - Degeneracy and sample impoverishment:} 
The cyclic evaluation of a finite set of particles (indicated in~Fig.~\ref{fig:ParticleAlg}) results in the problem of degeneracy which implies that many particles have negligible weights after a few iteration steps.
To address this issue, resampling methods have been introduced which remove the particles with low weights.
This leads to the so-called sample impoverishment, where the set of particles shrinks to a few members
with large weights.
Approaches tackling both problems of degeneracy and sample impoverishment can be categorized as follows:
In the first category, sophisticated resampling methods like dynamic resampling~\cite{zuo_dynamic_2013} or the use of particle-distribution optimization techniques~\cite{LI_2014} constitute an important step for making particle filters perform well in many practical scenarios~\cite{Branko_2004,Res2004,bolic2005,Xiaoyan2010}. However, these modifications of the classical particle filter are associated with a high computational load~\cite{bengtsson2008,PFpositioning,doucet_t2009}.
In the second category, methods like the Gaussian particle filter proposed in~\cite{GPF} approximate the posterior PDF as a continuous distribution, so that a dedicated sampling strategy inherently prevents the problems of degeneracy and sample impoverishment.\\[1mm]
\hspace*{3mm}\textit{Pr3 - Generalization:}
In addition to the described limitations, the intention of classical particle filtering (as initially proposed for tracking applications~\cite{Particle2}) has been to solve the instantaneous optimization problem without generalizing the instantaneous solution. This is effected by~(\ref{equ:weights}), where the particle weights are calculated considering only the observation likelihood $p( d_n| \mathbf{z}^{(l)}_n  )$ at time instant $n$.
This property of classical particle filtering is a disadvantage for the identification of high-dimensional systems that have a distinct static or slowly time-varying component. To give an example, assume that the entries of the state vector $\mathbf{z}_n$ to be modeled are random variables with time-invariant mean vector and time-variant uncertainty (e.g., modeled by additive Gaussian random variables of zero mean). The intention of classical particle filtering is to solve the instantaneous solution of the optimization problem without generalization to the global solution. Then, in particular, the resulting system estimate is not necessarily converging to the mean vector of the latent state vector (and thus not identifying an unbiased estimate for the unknown system). A more detailed description is part of Section~\ref{sec:CompHGM}.
\section{The \ac{EPFES} for nonlinear system identification}
\label{cha:EPFES}
In this section, we present the elitist particle filter based on evolutionary strategies (\ac{EPFES}) with the goal to overcome the previously described problems Pr1, Pr2 and Pr3 of classical particle filtering.
First, the properties of the \ac{EPFES} are explained in detail followed by an algorithmic realization and an overview on the optimization parameters.
Second, the \ac{EPFES} is shown (by proper choice of two parameters) to lead to the Gaussian particle filter. 
\subsection{Properties of the \ac{EPFES}}
\label{cha:EPFESproperties}
\begin{figure}[!t]
\input{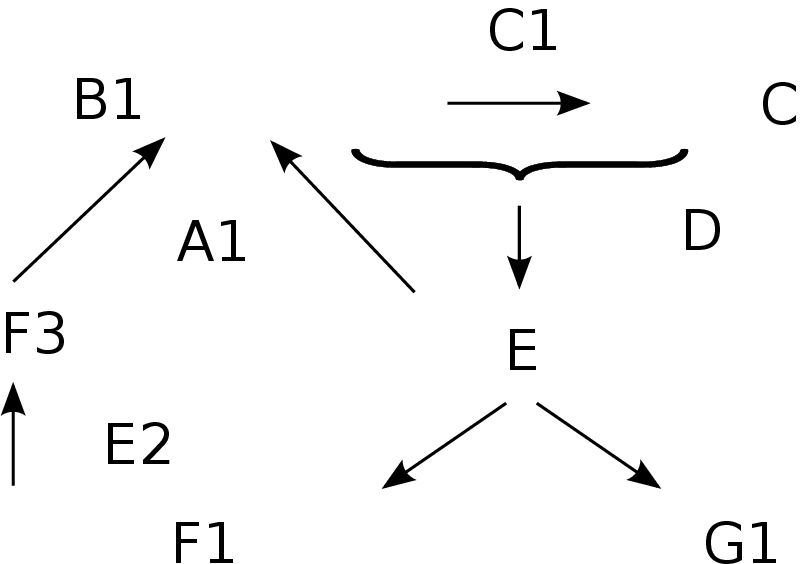}
\caption{Concept of the \ac{EPFES} with approximated posterior \ac{PDF}}
\label{fig:ParticleAlgNew}
\end{figure}

In this part, we introduce the properties S1, S2 and S3 of the EPFES motivated by the classical particle filter's problems Pr1, Pr2 and Pr3 (see previous section), respectively.\newpage
\noindent \hspace*{3mm}\textit{S1 - Selection of elitist particles:} Motivated by problem~Pr1, we define a selection process
associated with the so-called ``natural selection'' \cite{Evolution} following evolutionary strategies~(ES):
At time instant $n$, we consider the set of $L$ samples $\mathbf{z}^{(l)}_n$~and corresponding weights $\omega^{(l)}_n$.
In the selection process, the samples with a weight value smaller than a threshold $\omega_{\text{th},n}$ are dropped \cite{cho_2014}.
Due to its widespread use in evolutionary game theory, we propose to employ the weight average as adaptive threshold~\cite{Foster90,ma_dynamic_2011}
\begin{equation}
\omega_{\text{th},n} = \sum_{l=1}^L \omega_{n}^{(l)},
\label{equ:THmean}
\end{equation}
which is novel compared to our previous work ($\omega_{\text{th},n}$ as time-invariant tuning parameter~\cite{epfes}).
The remaining set of $Q_n\leq L$ selected samples represents the set of so-called elitist particles \cite{ParticleElite}~$\mathbf{z}^{(q)}_{\text{el},n}$ and is characterized by the normalized weights
\begin{equation}
\omega^{(q)}_{\text{el},n} = \frac{p\left( d_n|  \mathbf{z}^{(q)}_{\text{el},n}  \right)}{\sum\limits_{q=1}^{Q_n} p\left( d_n| \mathbf{z}^{(q)}_{\text{el},n}  \right)},
\label{equ:weightsNewest}
\end{equation}
where $q=1,...,Q_n$\footnote{Time dependencies of sample indices are omitted for simplicity.}.
Based on~(\ref{equ:ParticleFilUpdate1}), the estimate of the state vector is given as 
weighted superposition of the~elitist~particles
\begin{equation}
\mathbf{\hat{z}}_n = \sum_{q=1}^{Q_n} \omega^{(q)}_{\text{el},n}   \mathbf{z}^{(q)}_{\text{el},n}.
\label{equ:ParticleFilUpdateNew}
\end{equation}
\hspace*{3mm}\textit{S2 - Innovation by mutation:}
To cope with problem Pr2, the estimated state vector $\mathbf{\hat{z}}_n$ is employed to approximate the discrete distribution $p \left( \mathbf{z}_n | d_{1:n}\right)$
by a continuous \ac{PDF} $\hat{p} \left( \mathbf{z}_n | d_{1:n}\right) $, as proposed for the Gaussian case in \cite{GPF} and is discussed below (see Table~\ref{tab:GaussianPosterior}).
This has the advantage of addressing degeneracy and sample impoverishment without introducing resampling methods:
By drawing $R_n=L-Q_n$ samples $\mathbf{z}^{(r)}_{\text{in},n}$ ($r=1,...,R_n$) from the approximated distribution $\hat{p} \left( \mathbf{z}_n | d_{1:n}\right) $, we introduce innovation by refilling the set of particles.
In the terminology of evolutionary strategies~(ES), this introduction of innovation can be identified as mutation.
Finally, the time update is realized by drawing samples from $p(\mathbf{z}_{n+1}|\mathbf{z}_{n}=\mathbf{\bar{z}}^{(l)}_{n})$ \cite{GPF}, where $\mathbf{\bar{z}}_n^{(l)}$ denotes the refilled set of samples.
Based on the process equation in~(\ref{equ:TranEq}), this leads~to~\cite{GPF}
\begin{equation}
\mathbf{z}^{(l)}_{n+1} = \mathbf{f} \left(\mathbf{\bar{z}}^{(l)}_{n} \right) + \mathbf{w}^{(l)}_n \quad \text{for} \quad l=1,...,L
\label{equ:TranEqIns}
\end{equation}
with samples $\mathbf{w}^{(l)}_n$ drawn from the \ac{PDF} $p(\mathbf{w}_n)$.\\[1mm]
\hspace*{3mm}\textit{S3 - Long-term fitness measures:} The evaluation of the particles following~(\ref{equ:weights}) yields an instantaneous estimate of the particle weights.
As discussed in problem Pr3, it might be of interest for the identification of high-dimensional systems with slowly time-varying components to incorporate information about the previous time instants into the particle evaluation as well. For this, we 
employ the following weight calculation based on the assumption of a normally distributed observation uncertainty $v_n$ in~(\ref{equ:Gaussv}) (detailed derivation in the appendix):
\begin{align} &\omega_n^{(l)}= 
  \frac{  \text{exp} \left(\hat{\omega}_n^{(l)} \right) }{\sum\limits_{l=1}^L  \text{exp} \left(\hat{\omega}_n^{(l)} \right) }, \quad  \hat{\omega}_n^{(l)}=\lambda \hat{\omega}_{n-1}^{(l)} + (1-\lambda) \tilde{\omega}_n^{(l)}. 
  \label{equ:weightsNew}
  \end{align}
Therein, we estimate the instantaneous update as
\begin{align}
\tilde{\omega}_n^{(l)}= \frac{(d_{n} - \mathbf{g} \{\mathbf{z}^{(l)}\})^2}{c_{\sigma,n}}, \quad  c_{\sigma,n} = -2\sigma_{v,n}^2 \frac{1-\lambda}{1+\lambda},
  \label{equ:weightsInnov}
\end{align}
initialize $\hat{\omega}_{1}^{(l)}=\tilde{\omega}_{1}^{(l)}$ and thus define $\hat{\omega}_{n-1}^{(l)}=\tilde{\omega}_{n-1}^{(l)}$ for set of non-elitist particles $\mathbf{z}^{(r)}_{\text{in},n-1}$ drawn at the time instant $n-1$.\\[1mm]
An overview of the \ac{EPFES} is given in Fig.~\ref{fig:ParticleAlgNew}.
In comparison to the classical particle-filter concept in Fig.~\ref{fig:ParticleAlg}, the discrete distribution is replaced by an approximated continuous \ac{PDF} $\hat{p} \left( \mathbf{z}_n | d_{1:n}\right)$ and the evolutionary selection process is integrated, which
facilitates the introduction of innovation into the set of particles
by taking samples from $\hat{p} \left( \mathbf{z}_n | d_{1:n}\right)$.
Note that this evolutionary selection process intends to capture dynamic nonlinearities
by mutation and time-invariant system components by the evolutionary selection with recursively-calculated particle weights.
\subsection{Realization of the EPFES}
\label{sec:RealEPFES}
In this section, we illustrate the practical implementation of the EPFES by considering the overview shown in Table~\ref{tab:EPFESalg}.
We start with a finite set of $L$ samples $\mathbf{z}^{(l)}_{n}$ and respective values $\hat{\omega}^{(l)}_{n-1}$ estimated in~(\ref{equ:weightsNew}).
First, the particle weights are recursively updated by inserting (\ref{equ:weightsInnov}) into (\ref{equ:weightsNew}).
Then, we select $Q_n$ elitist particles $\mathbf{z}^{(q)}_{\text{el},n}$ with weights larger than~$\omega_{\text{th},n}$ in (\ref{equ:THmean}) and consider the following distinction for estimating the posterior PDF  $\hat{p} \left( \mathbf{z}_n | d_{1:n}\right)$:
If no elitist particles have been selected ($Q_n=0$), the entire set of (non-elitist) particles and respective weights $\{\mathbf{z}^{(l)}_{n},\omega^{(l)}_{n}\}$ is employed to approximate
the posterior PDF $\hat{p} \left( \mathbf{z}_n | d_{1:n}\right)$ with mean vector $\mathbf{\hat{z}}_n$ calculated according to (\ref{equ:ParticleFilUpdate1}).
In case of $Q_n>0$, we estimate the posterior PDF $\hat{p} \left( \mathbf{z}_n | d_{1:n}\right)$
using the set of $Q_n$ elitist particles and respective weights $\{\mathbf{z}^{(q)}_{\text{el},n}, \omega^{(q)}_{\text{el},n}\}$, where the mean vector $\mathbf{\hat{z}}_n$ is determined according to~(\ref{equ:ParticleFilUpdateNew}).
To give an example for the estimation of the posterior PDF, consider the special case of $\hat{p} \left( \mathbf{z}_n | d_{1:n}\right) \approx \mathcal{N} \{ \mathbf{\hat{z}}_n, \mathbf{\hat{C}}_{\mathbf{z},n} \}$ shown in Table~\ref{tab:GaussianPosterior}.
Finally, the $R_n$ non-elitist particles are replaced by samples drawn from the posterior~\ac{PDF} $\hat{p} \left( \mathbf{z}_n | d_{1:n}\right)$ and the time update realized following~(\ref{equ:TranEqIns}).\clearpage
\begin{table}[!t]
\caption{Overview on the implementation steps of the EPFES}
\vspace{-2mm}
\begin{tabular}[c]{l}
\hline\\[-6mm]
\textbf{Starting point}: $L$ samples $\mathbf{z}^{(l)}_n$ and estimates $\hat{\omega}^{(l)}_{n-1}$ \\[0mm]
\textbf{Step 1: Calculate particle weights} $\omega^{(l)}_{n}$ according to (\ref{equ:weightsNew}) \\[0mm]
\textbf{Step 2: Select $Q_n$ elitist particles} with weights larger than $ \omega_{\text{th},n}$ in (\ref{equ:THmean}) \\[0mm]
\textbf{Step 3: Estimate posterior PDF $\hat{p} \left( \mathbf{z}_n | d_{1:n}\right)$ with mean vector $\mathbf{\hat{z}}_n$} \\[-2mm]
\quad if $Q_n>0$\\[-1mm]
\quad \qquad $-$ Normalize weights of elitist particles according to~(\ref{equ:weightsNewest})\\[-2mm]
\quad \qquad $-$ Use elitist particles to estimate $\hat{p} \left( \mathbf{z}_n | d_{1:n}\right)$\\[-4mm]
\quad else\\[-4mm]
\quad \qquad $-$ Use all particles to estimate $\hat{p} \left( \mathbf{z}_n | d_{1:n}\right)$\\[-4mm]
\quad end\\[-1mm]
\textbf{Step 4: Introduce innovation} \\[-2mm]
\quad \qquad $-$ Replace non-elitist particles by samples from $\hat{p} \left( \mathbf{z}_n | d_{1:n}\right)$\\[-2mm]
\quad \qquad $-$ Calculate $\hat{\omega}_{\text{th},n}$ = $\tilde{\omega}_{\text{th},n}$ for the new samples following (\ref{equ:weightsInnov}) \\[0mm]
\textbf{Step 5:} \textbf{Determine $\mathbf{z}^{(l)}_{n+1}$} using the time update of (\ref{equ:TranEqIns}) $\rightarrow n=n+1$\\[1mm]
\hline
\end{tabular}
\label{tab:EPFESalg}
\end{table}
\noindent Note that in principle, the realization of the EPFES is not restricted to approximating the posterior PDF by a Gaussian probability distribution. For instance, a uniform posterior PDF has been shown in~\cite{epfes} to be promising for the task of nonlinear~\ac{AEC}.
However, in this article we focus on a Gaussian probability distribution as approximation for the posterior PDF to directly establish the link between the EPFES and the Gaussian particle filter in the following section. 
%
\subsection{Gaussian particle filter as a simple special case}
\label{sec:CompEpfesGPF}
The Gaussian particle filter has been applied to a variety of applications (see e.g. \mbox{\cite{wu_g2007,bolic_s2010,Lim2013,Rao2013}}) and conceptionally differs from classical particle filtering (reviewed in Section~\ref{sec:SSmodel}) by approximating the posterior PDF as Gaussian probability distribution~\cite{GPF}.
The following assumptions A1 and A2 simplify the EPFES to the Gaussian particle filter proposed~in~\cite{GPF}:\\[1mm]
\hspace*{3mm}\textit{A1 - Gaussian posterior PDF:}
The approximated posterior \ac{PDF} is assumed to be a Gaussian probability distribution
 \begin{equation}
\hat{p} \left( \mathbf{z}_n | d_{1:n}\right) \approx \mathcal{N} \{ \mathbf{\hat{z}}_n, \mathbf{\hat{C}}_{\mathbf{z},n} \},  
\label{equ:PosteriorGauss}
 \end{equation}
where we estimate the mean vector $\mathbf{\hat{z}}_n$ and the covariance matrix $\mathbf{\hat{C}}_{\mathbf{z},n}$ as summarized in Table~\ref{tab:GaussianPosterior} (see Section~\ref{sec:RealEPFES} for more details about the realization of the EPFES). \\[1mm]
\hspace*{3mm}\textit{A2 - No elitist particles ($Q_n=0$):}
The evolutionary selection mechanism is deactivated by setting the threshold~to
 \begin{equation}
\omega_{\text{th},n}=1.
 \end{equation}
This leads to $Q_n=0$ (no~elitist particles are selected) and the approximation of
the posterior \ac{PDF} $\hat{p} \left( \mathbf{z}_n | d_{1:n}\right)$ using the entire set of $L$ particles and its respective weights. As shown in~Table~\ref{tab:GaussianPosterior},
all $L$ particles $\mathbf{z}^{(l)}_n$ are subsequently replaced by new samples drawn from the approximated \ac{PDF} $\hat{p} \left( \mathbf{z}_n | d_{1:n}\right)$.
This implies that for the Gaussian particle filter it is not possible to evaluate particles based on long-term fitness measures.\\
Besides the choice of a Gaussian posterior PDF in~(\ref{equ:PosteriorGauss}), the major conceptional extension of the EPFES with respect to the Gaussian particle filter is the evolutionary selection of elitist particles which is active for \mbox{$0\leq\omega_{\text{th},n}<1$}.
This evolutionary choice of elitist particles provides the opportunity to evaluate samples based on long-term fitness measures.
To this end, the smoothing factor $\lambda$~in~(\ref{equ:weightsNew}) can be chosen dependent on the specific scenario.
The impact of the evolutionary selection process on the system identification performance of the EPFES will be investigated for two different scenarios in Section~\ref{cha:Experiments}.
\begin{table}[!t]
\caption{Sampling as part of the EPFES with approximated posterior PDF $\hat{p} \left( \mathbf{z}_n | d_{1:n}\right) \approx \mathcal{N} \{ \mathbf{\hat{z}}_n, \mathbf{\hat{C}}_{\mathbf{z},n} \}$,
where we differentiate with respect to the number of $Q_n$ elitist particles}
\centering
\vspace{-1mm}
\begin{tabular}[c]{|c|c|}
\hline
\multirow{4}{*}{\vspace*{-8mm}$Q_n>0$}
		& \specialcell{}\\[-4mm]
		& \specialcell{$\mathbf{\hat{z}}_n = \sum\limits_{q=1}^{Q_n} \omega^{(q)}_{\text{el},n}   \mathbf{z}^{(q_n)}_{\text{el},n}$}\\[2mm]
		& \specialcell{$\mathbf{\hat{C}}_{\mathbf{z},n} = \frac{1}{Q_n} \sum\limits_{q=1}^{Q_n}(\mathbf{\hat{z}}_n - \mathbf{z}^{(q)}_{\text{el},n})(\mathbf{\hat{z}}_n - \mathbf{z}^{(q)}_{\text{el},n})^\text{T}$}\\[0mm]
		& \specialcell{$\rightarrow$ Draw $L-Q_n$ samples from $\mathcal{N} \{ \mathbf{\hat{z}}_n, \mathbf{\hat{C}}_{\mathbf{z},n} \}$}\\[2mm]
		\hline
\multirow{4}{*}{\vspace*{-7mm}$Q_n=0$}
		& \specialcell{}\\[-4mm]
		& \specialcell{$\mathbf{\hat{z}}_n = \sum\limits_{l=1}^{L} \omega^{(l)}_n   \mathbf{z}^{(l)}_n$}\\[2mm]
		& \specialcell{$\mathbf{\hat{C}}_{\mathbf{z},n} = \frac{1}{L} \sum\limits_{l=1}^{L}(\mathbf{\hat{z}}_n - \mathbf{z}^{(l)}_{n})(\mathbf{\hat{z}}_n - \mathbf{z}^{(l)}_{n})^\text{T}$}\\[0mm]
		& \specialcell{$\rightarrow$ Draw $L$ samples from $\mathcal{N} \{ \mathbf{\hat{z}}_n, \mathbf{\hat{C}}_{\mathbf{z},n} \}$}\\[2mm]
		\cline{1-2}
\end{tabular}\vspace*{-2mm}
\label{tab:GaussianPosterior}
\end{table}
\section{Experimental results}
\label{cha:Experiments}
In this section, we focus on evaluating the system identification performance of the \ac{EPFES}.
In particular, we follow the assumption A1 of the previous section and consider a Gaussian posterior PDF defined in~(\ref{equ:PosteriorGauss}) with parameter estimation summarized in Table~\ref{tab:GaussianPosterior}. Thus, the
commonly employed Gaussian particle filter is represented by the special case of $\omega_{\text{th},n} = 1$ and employed as reference algorithm to evaluate the system identification improvement using the evolutionary elitist-particle selection process.\\
\begin{table}[!b]
\caption{Overview of both tasks including the structures of process and observation equations}
\centering
\vspace{-1mm}
\begin{tabular}[c]{|c|c|c|}
\hline
 & Task~A, Section~\ref{sec:CompGPF} & Task~B, Section~\ref{sec:CompHGM} \\
\hline
Process equation & $z_n = f \left(z_{n-1} \right) + w_n$ & $\mathbf{z}_n =\mathbf{z}_{n-1} + \mathbf{w}_n$ \\
Observation equation & $d_n = g \left(z_{n} \right) + v_n$ & $d_n = \mathbf{g} \left( \mathbf{z}_{n}\right) + v_n$ \\
\hline
\end{tabular}
\label{tab:Overexp}
\end{table}
The evaluation is split into two parts, where
the respective structures of process and observation equations are described in~Table~\ref{tab:Overexp}: 
Task~A represents a widely-used application of classical particle filtering, where a latent state variable $z_n$ is quickly changing over time with an additive uncertainty $w_n$ of large variance. This removes the need for
employing long-term fitness measures, so that the \ac{EPFES} is realized by setting the smoothing factor $\lambda$ in (\ref{equ:weightsNew}) equal~to~0.
In Task~B, we model the temporal progress of a slowly-varying latent state vector $\mathbf{z}_n$ by a linear process equation including an additive uncertainty~$\mathbf{w}_n$.
As a result of this, we apply the EPFES to verify the generalization properties of the instantaneous solution of the optimization problem dependent on the smoothing factor $\lambda$ in the weight update of~(\ref{equ:weightsNew}).
To this end, we modify the length of the vector $\mathbf{z}_n$ to evaluate the system identification performance of the EPFES for different sizes~of~the~search~space.
\subsection{Univariate Nonstationary Growth Model (UNGM)}
\label{sec:CompGPF}
\begin{figure}[!b]
 \centering
  \begin{tikzpicture}[scale=1]
\begin{axis}[
     width=8.8cm,height=3.5cm,grid=major,grid style = {dotted,black},
     ylabel style={yshift=-1mm},
     xlabel style={xshift=-1mm},
     ylabel={$z_n$ $\; \rightarrow$},
     xlabel={$n$ $\; \rightarrow$},
     ymin=-21, ymax=21,xmin=1,xmax=100,extra x ticks={0,100},extra y ticks={-20,20},
     legend style={nodes=right},
 ]
	\addplot[thin,blue,solid] table [x index=0, y index=1]{EPFES_x.dat};
  \end{axis}
 	\end{tikzpicture}\vspace*{-1.8mm}
\caption{Realization of the process equation in (\ref{equ:ProcUNG}) with the parameter settings of (\ref{equ:ParUNGM}) for $N=100$ time instants.}\vspace*{-2mm}
\label{fig:ProcUNGM}
\end{figure}
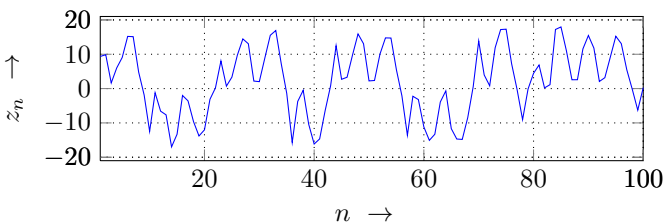
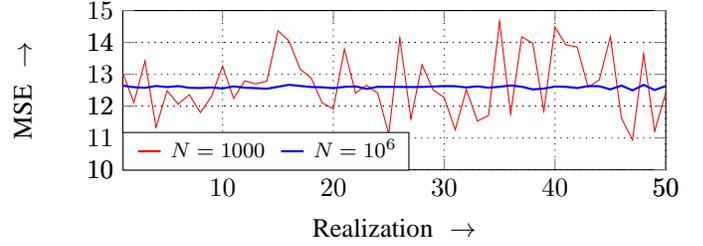
\begin{figure}[!t]
 \centering
  \begin{tikzpicture}[scale=1]
  \def\lx{0}
	\def\ly{0}
	\begin{axis}[
     width=8.8cm,height=3.7cm,grid=major,grid style = {dotted,black},
     ylabel style={yshift=1mm},
     ylabel={MSE $\; \rightarrow$},
     xlabel={Realization $\; \rightarrow$},
     ymin=10, ymax=15,xmin=1,xmax=50,extra x ticks={0,50},extra y ticks={10,11,12,13,14,15},
     legend style={nodes=right},
 ]
	\addplot[thin,red,solid] table [x index=0, y index=1]{MSE_gpf1000.dat};
	\addplot[thick,blue,solid] table [x index=0, y index=1]{MSE_gpf1e6.dat};
  \end{axis}
  \draw[fill=white] (\lx,\ly) rectangle (3.8+\lx,0.5+\ly);
 	\draw[red,thick,solid] (0.2+\lx,0.25+\ly) -- +(0.3,0) node[anchor=mid west,black] {\footnotesize $N=1000$};
	\draw[blue,thick,solid] (2.1+\lx,0.25+\ly) -- +(0.3,0) node[anchor=mid west,black] {\footnotesize $N=10^6$};
  \end{tikzpicture}\vspace*{-2mm}
\caption{Simulation results for the MSE defined in~(\ref{equ:MSE}) with $K=1$, where we realize the same experiment $50$ times to evaluate the impact of choosing $N=1000$ or $N=10^6$ iteration steps on the variations of the MSE.}
\label{fig:VarGPF}	
\end{figure}
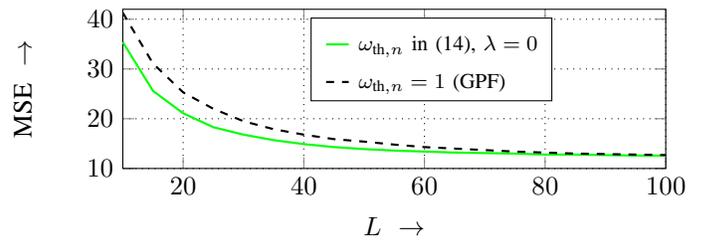
\begin{figure}[!b]
 \centering
  \begin{tikzpicture}[scale=1]
  \def\lx{2.5}
	\def\ly{0.9}
 	\begin{axis}[
     width=8.8cm,height=3.7cm,grid=major,grid style = {dotted,black},
     ylabel style={yshift=1mm},
     ylabel={MSE $\; \rightarrow$},
     xlabel={$L $ $\; \rightarrow$},
     ymin=10, ymax=42,xmin=10,xmax=100,
     legend style={nodes=right},
 ]
	\addplot[thick,green,solid] table [x index=0, y index=1]{MSE.dat};
  \addplot[thick,black,dashed] table [x index=0, y index=1]{MSE_GPF.dat};
  \end{axis}
         \draw[fill=white] (\lx,\ly) rectangle (3.2+\lx,1.1+\ly);
	\draw[green,thick,solid] (0.2+\lx,0.75+\ly) -- +(0.3,0) node[anchor=mid west,black] {\footnotesize  $\omega_{\text{th},n}$ in~(\ref{equ:THmean}), $\lambda=0$};
	\draw[black,thick,dashed] (0.2+\lx,0.25+\ly) -- +(0.3,0) node[anchor=mid west,black] {\footnotesize $\omega_{\text{th},n}=1$ (GPF)};
 	\end{tikzpicture}\vspace*{-2mm}
\caption{Simulation results for the MSE defined in~(\ref{equ:MSE}) for two different choices of estimating the parameter $\omega_{\text{th},n}$, where $\omega_{\text{th},n}=1$ represents the Gaussian particle filter (GPF), \mbox{$N=10^6$} and $K=50$.}
\label{fig:elitesEPFES}
\end{figure}

In order to evaluate the elitist-particle selection of the EPFES without recursively calculated particle weights, we consider a frequently employed benchmark system, known as \ac{UNGM}.
The \ac{UNGM} has been employed in~\cite{GPF} to compare the system identification performance of the Gaussian particle filter with the unscented Kalman filter, the extended Kalman filter and sequential importance sampling with resampling.\\
\noindent The \ac{UNGM} is described by the observation equation
\begin{equation}
d_n =  \frac{z_{n}^2 }{20} + v_n \quad \text{with} \quad  v_n \sim \mathcal{N} \left\{0, 1 \right\} 
\label{equ:ObsUNG}
\end{equation}
\noindent relating $d_n$ to the latent state variable $z_n$ by a quadratic \mbox{function} and a normally distributed observation uncertainty~$v_n$.
The \ac{UNGM}'s temporal progress of the state variable is modeled by the process equation:
\begin{equation}
z_n = \alpha z_{n-1} + \beta \frac{z_{n-1}}{1+z_{n-1}^2 } + \gamma \cos (1.2(n-1)) + w_n,
\label{equ:ProcUNG}
\end{equation}
which is a nonlinear time-variant function dependent on the parameters $\alpha$, $\beta$, $\gamma$ as well as on the additive uncertainty $w_n$ chosen as follows~\cite{GPF}:
\begin{equation}
 w_n \sim \mathcal{N} \left\{0, 1 \right\}, \; \alpha=0.5, \; \beta=25,\; \gamma=8.
 \label{equ:ParUNGM}
\end{equation}
To emphasize the time-variant characteristics of the latent state variable, one realization of the process equation for $N=100$ time instants is illustrated in Fig.~\ref{fig:ProcUNGM}.\\
\noindent As a measure to evaluate the system identification performance, we use the mean square error
\begin{equation}
\text{MSE} = \frac{1}{KN} \sum\limits_{k=1}^K \sum\limits_{n=1}^N (z_n-\hat{z}_{n,k})^2
\label{equ:MSE}
\end{equation}
\noindent between the state variable $z_n$ and its estimate $\hat{z}_{n,k}$, averaged over $K$ realizations of the same experiment.
This is motivated by the variance in the system identification performance of sequential Monte Carlo methods like the Gaussian particle filter due to the numerical sampling from a continuous \ac{PDF}~\cite{GPF}. 
As shown for the Gaussian particle filter in Fig.~\ref{fig:VarGPF}, the resulting variations in evaluation measures like the MSE can be reduced by taking long-term averages, where we choose $N=10^6$ and $K=50$ in the following.
\\
\noindent The parameters of the \ac{EPFES} are chosen as follows:
\begin{itemize}
 \item The particle weights in~(\ref{equ:weightsNew}) are calculated in a non-recursive way by setting $\lambda=0$. This is motivated by the strongly time-variant observation and process equation combined with 
additive uncertainties $v_n$ and $w_n$ of large variance, cf.~(\ref{equ:ObsUNG}) and~(\ref{equ:ParUNGM}). The resulting oscillations in the temporal evolution of both state variable $z_n$ and observation $d_n$ removes the need for using long-term fitness measures.
 \item The threshold $\omega_{\text{th},n}$ is estimated following~(\ref{equ:THmean}). To realize the Gaussian particle filter as reference algorithm, the threshold is set to $\omega_{\text{th},n}=1$. 
\end{itemize}
The simulation results for the MSE defined~in~(\ref{equ:MSE}) are shown in Fig.~\ref{fig:elitesEPFES} in dependence of the number of $L$ particles.
Besides the well-known tendency that the system identification performance increases for higher values of $L$, we experience that the EPFES achieves lower MSE scores compared to the Gaussian particle filter.
This performance gain of the EPFES reduces with increasing number of $L$ and is most pronounced for the practically-relevant case of a small set of $L$ particles (see MSE values in Table~\ref{tab:OverEPFES}).
Note that the computational complexity of the EPFES is comparable with respect to the Gaussian particle filter, as the additional number of real-valued multiplications to calculate the threshold in~(\ref{equ:THmean}) and to normalize the elitist weights in (\ref{equ:weightsNewest}) is almost compensated by the reduced number of elitist particles to calculate the \ac{MMSE} estimate in~(\ref{equ:ParticleFilUpdateNew}).\newline
\begin{table}[!t]
\caption{MSE of the EPFES and the Gaussian particle filter (GPF) according to~(\ref{equ:MSE}) dependent on the number of $L$ particles.}
\centering
\begin{tabular}[c]{|c|c|c|c|c|}
\hline
Threshold & $L=10$ & $L=20$ & $L=50$ & $L=100$ \\
\hline
$\omega_{\text{th},n}$ in~(\ref{equ:THmean}) & 35.4 & 21.2 & 13.9 & 12.6 \\
\hline
$\omega_{\text{th},n}=1$ (GPF) & 41.3 & 31.1 & 15.4 & 12.7 \\
\hline
\end{tabular}
\label{tab:OverEPFES}
\end{table}
\textit{Summary of Task A:} We evaluated the system identification performance of the EPFES using the \ac{UNGM} as benchmark system with strongly time-variant observation and process equation.
By setting the smoothing factor $\lambda$ to zero, it is shown that, in comparison with the Gaussian particle filter, the system identification improvement achieved by the EPFES is most prominent for a small number of
$L$ particles (see Table~\ref{tab:OverEPFES}).
\subsection{Nonlinear Acoustic Echo Cancellation (AEC)}
\label{sec:CompHGM}
\begin{figure}[!b]
\centering
\includegraphics[width=0.44\textwidth]{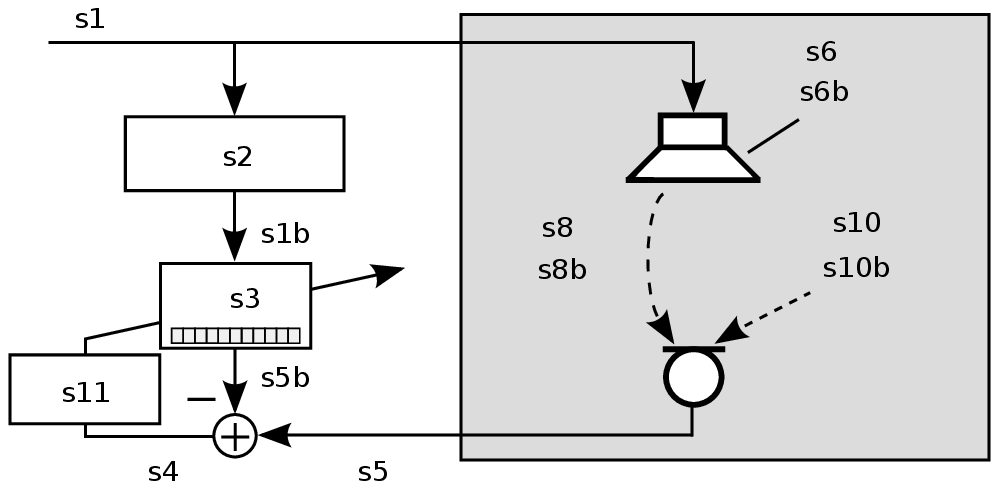}%
\caption{The NL-AEC scenario with memoryless preprocessor and linear FIR filter $\mathbf{\hat{h}}_n$, where $\mathbf{\hat{h}}_n$ is estimated by the NLMS algorithm according to~(\ref{equ:NLMS}).}
\label{fig:NONlinAEC}
\vspace{-3mm}
\end{figure}

In this section, we consider the scenario for nonlinear \ac{AEC} shown in Fig.~\ref{fig:NONlinAEC} as a real-world example for the identification of a slowly time-varying nonlinear system. The goal is to identify the electro-acoustic echo path (from the loudspeaker to the microphone) for simultaneously active near-end interferences.
Here, nonlinear loudspeaker signal distortions created by transducers and amplifiers in minituarized loudspeakers reduce the performance of linear echo path models.
As such nonlinearly distorting loudspeakers are followed by a linear transmission path through air, modeling the overall echo path as a nonlinear-linear cascade is a reasonable approximation~\cite{Stenger}.
The structure of this Hammerstein system is shown in the left half of Fig.~\ref{fig:NONlinAEC}, where a memoryless preprocessor realizes an element-wise transformation of the input vector
\begin{equation}
\mathbf{x}_n=[x_n,x_{n-1},...,x_{n-M+1}]^\text{T}
\label{equ:x}
\end{equation}
(with time-domain samples $x_n$) to the output vector $\mathbf{y}_n$ of equal length.
Motivated by the good performance in nonlinear \ac{AEC}~\cite{Hofmann,Huemmer2014,saepfes}, we choose a polynomial preprocessor based on odd-order Legendre functions of the first~kind:
\begin{equation}
\mathbf{y}_n=\mathbf{f}(\mathbf{x}_n,\mathbf{\hat{a}}_{n}) =  \mathbf{x}_n + \sum_{\nu=1}^{3} \hat{a}_{\nu,n} \cdot \mathcal{L}_{2\nu+1} \{\mathbf{x}_n\},
\label{equ:y}
\end{equation}
which is parameterized by the estimated vector
\begin{equation}
\mathbf{\hat{a}}_n=[\hat{a}_{1,n},\hat{a}_{2,n},\hat{a}_{3,n}]^\text{T}
\label{equ:aN}
\end{equation}
with coefficients $\hat{a}_{\nu,n}$, where $\nu=1,2,3$.
%
%
%
\psfrag{s1}[c][c]{$\mathbf{x}_n$}%
\psfrag{s1b}[c][c]{$\mathbf{y}_n$}%
\psfrag{s2}[c][c]{$\mathbf{f}(\mathbf{x}_n,\mathbf{\hat{a}}_n)$}%
\psfrag{s3}[c][c]{$\mathbf{\hat{h}}_n$ }%
\psfrag{s4}[c][c]{$e_n$ }%
\psfrag{s5}[c][c]{$d_n$ }%
\psfrag{s5b}[c][b]{$\hat{d}_{n}$}%
\psfrag{s6}[c][c]{\small Non-}%
\psfrag{s6b}[c][c]{\small linearities}%
\psfrag{s7}[c][c]{$\mathbf{f}(\mathbf{x}_n,\mathbf{a}_n)$}%
\psfrag{s8}[c][c]{\small Acoustic}%
\psfrag{s8b}[c][c]{\small path}%
\psfrag{s9}[c][c]{\small Uncertainty}%
\psfrag{s10}[c][c]{\small Near-end}%
\psfrag{s10b}[c][c]{\small interferences}%
\psfrag{s11}[c][c]{\textcolor{black}{NLMS}}%
%
%

\begin{figure}[!t]
\centering
\includegraphics[width=0.48\textwidth]{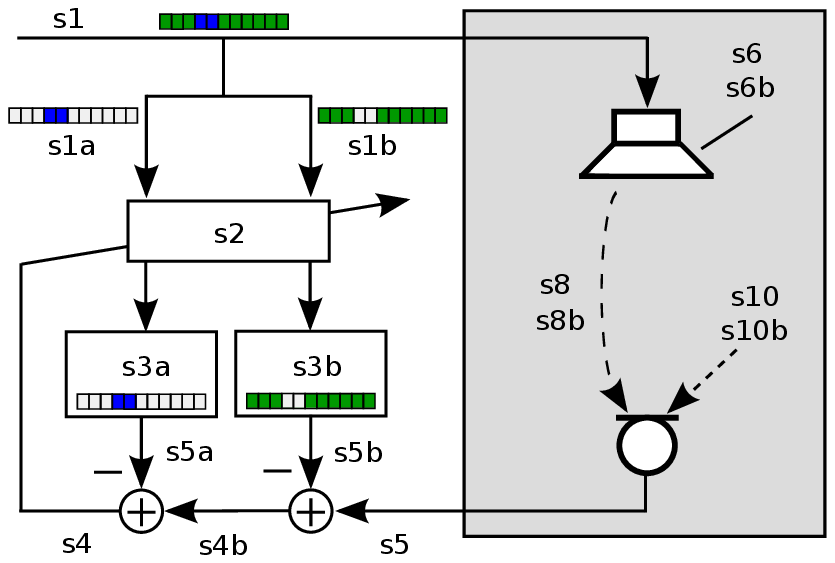}%
\caption{The NL-AEC scenario with memoryless preprocessor and linear FIR filter, where the parameter vector $\mathbf{\hat{a}}_n$ is estimated using the direct-path microphone signal component $d_{\text{c},n}$.}
\label{fig:NONlinAECsaf}
\vspace{-4mm}
\end{figure}

The acoustic path at time $n$ between loudspeaker and microphone is estimated by the linear \ac{FIR} filter
\begin{equation}
\mathbf{\hat{h}}_n=[\hat{h}_{0,n},\hat{h}_{1,n},...,\hat{h}_{M-1,n}]^\text{T}
\label{equ:hGen}
\end{equation}\\
using the \ac{NLMS} algorithm:
\begin{equation}
\mathbf{\hat{h}}_{n+1}= \mathbf{\hat{h}}_{n} +\frac{ \mu }{\mathbf{y}^\text{T}_n\mathbf{y}_n + \epsilon} \;\mathbf{y}_n e_n,
\label{equ:NLMS}
\end{equation}
with the scalar stepsize $\mu$, a positive constant $\epsilon$ to avoid division by zero and the error signal
$e_n = d_n-\hat{d}_n$
relating the observation $d_n$ and its estimate $ \hat{d}_n=\mathbf{\hat{h}}_{n}^\text{T}  \mathbf{y}_n$~\cite{AdaptiveFilter}.\\
It should be emphasized that the adaptation of a Hammerstein system is highly appropriate for investigating the generalization properties of the \ac{EPFES}:
The adaptation of the linear subsystem $\mathbf{\hat{h}}_n$ in (\ref{equ:NLMS}) depends on the preprocessor output $\mathbf{y}_n$ and thus strongly depends on the performance of the \ac{EPFES}, estimating the preprocessor coefficients $\mathbf{\hat{a}}_{n+1}$.\\
For the identification of the memoryless preprocessor, we apply the concept of \ac{SA} filtering initially proposed in~\cite{Hofmann}:
\begin{figure}[!b]
\centering
\begin{tikzpicture}[scale=1]
\def\lx{1.8}
\def\ly{1.2}
\begin{axis}[
      width=8.8cm,height=3.8cm,grid=major,grid style = {dotted,black},
      ylabel style={yshift=-2mm},
      ylabel={$\hat{h}_{\kappa,n}  $ $\; \rightarrow$},
      xlabel={tap $\kappa$ $\; \rightarrow$},
      ymin=-0.5, ymax=1,xmin=0,xmax=60,
      ]
      \addplot[thick,dunkelgruen,dashed] table [x index=0, y index=1]{hR1.dat};
      \addplot[thick,dunkelgruen,dashed] table [x index=0, y index=1]{hR2.dat};
      \addplot[thick,dunkelblau,solid] table [x index=0, y index=1]{hC.dat};
 \end{axis}
 \draw[fill=white] (\lx,\ly) rectangle (2.6+\lx,0.62+\ly);
 \draw[green,dashed,thick] (1.35+\lx,0.25+\ly) -- +(0.3,0) node[anchor=mid west,black] {\footnotesize $\mathbf{\hat{h}}_{\text{c},n}$};
 \draw[blue,thick,solid] (0.14+\lx,0.25+\ly) -- +(0.3,0) node[anchor=mid west,black] {\footnotesize $\mathbf{\hat{h}}_{\text{p},n}$};
 \end{tikzpicture}
 \caption{Example for splitting $\mathbf{\hat{h}}_n $
into the \textcolor{black}{direct-path peak region} $\mathbf{\hat{h}}_{\text{p},n}$
and the \textcolor{black}{complementary part} $\mathbf{\hat{h}}_{\text{c},n}$ with $I=12$, $M_\text{p}=5$ and $M=61$.} 
  \label{fig:ResultsSAF}
  \vspace{-4mm}
 \end{figure}
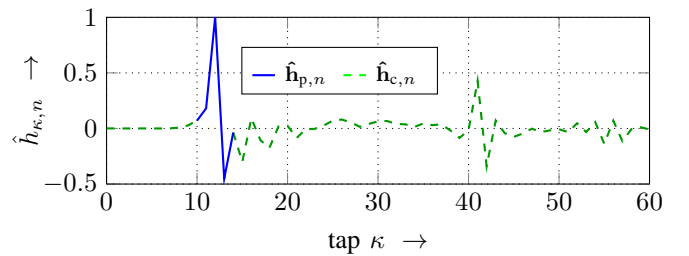
The linear subsystem (represented by the estimated \ac{RIR} vector $\mathbf{\hat{h}}_{n}$) is split into the
\textcolor{black}{direct-path peak region}
\begin{equation}
\mathbf{\hat{h}}_{\text{p},n}=[\hat{h}_{I-T,n},...,\hat{h}_{I+T,n}]^\text{T}
\label{equ:hC}
\end{equation}
and the \textcolor{black}{complementary part}
\begin{equation}
\mathbf{\hat{h}}_{\text{c},n}= [\hat{h}_{0,n},...,\hat{h}_{I-T-1,n},\hat{h}_{I+T+1,n},...,\hat{h}_{M-1,n}]^\text{T},
\label{equ:hR}
\end{equation}
based on the time lag $I$ of the energy peak in $\mathbf{\hat{h}}_{n}$ \cite{Hofmann}, where ${I \geq T}$. 
As exemplarily shown in Fig.~\ref{fig:ResultsSAF}, the direct-path part $\mathbf{\hat{h}}_{\text{p},n}$ is of length
$M_\text{p}=2T+1<M$ and the remaining part $\mathbf{\hat{h}}_{\text{c},n}$ is of length $M-M_\text{p}$. 
As indicated in Fig.~\ref{fig:NONlinAECsaf}, this concept of splitting $\mathbf{\hat{h}}_{n}$ into the two components $\mathbf{\hat{h}}_{\text{p},n}$ and $\mathbf{\hat{h}}_{\text{c},n}$ is similarily applied to the input vector $\mathbf{x}_n$ to define the signal vectors $\mathbf{x}_{\text{p},n}$ and $\mathbf{x}_{\text{c},n}$, respectively~\cite{saepfes}.
As a consequence of this, the direct-path microphone signal component
\begin{equation}
d_{\text{p},n} = d_n - \hat{d}_{\text{c},n}  =d_n - \mathbf{\hat{h}}_{\text{c},n}^\text{T} \mathbf{f}\left(\mathbf{x}_{\text{c},n},\mathbf{\hat{a}}_n \right)
\label{equ:ErrorSAF}
\end{equation}
can be introduced as an observed random variable which depends only on the direct-path component of the nonlinear echo path,~$\mathbf{\hat{h}}_{\text{p},n}$, 
because the influence of the remaining part $\mathbf{\hat{h}}_{\text{c},n}$ is assumed to be removed via (\ref{equ:ErrorSAF}).
This leads to the NL-AEC scenario shown in Fig.~\ref{fig:NONlinAECsaf}, where we estimate the preprocessor coefficients based on the error signal
\begin{equation}
e_n = d_{\text{p},n} - \hat{d}_{\text{p},n}  =d_{\text{p},n} - \mathbf{\hat{h}}_{\text{p},n}^\text{T} \mathbf{f}\left(\mathbf{x}_{\text{p},n},\mathbf{\hat{a}}_n \right)
\end{equation}
between the direct-path microphone signal component $d_{\text{p},n}$ and its estimate $\hat{d}_{\text{p},n}=\mathbf{\hat{h}}_{\text{p},n}^\text{T} \mathbf{f}\left(\mathbf{x}_{\text{p},n},\mathbf{\hat{a}}_n \right)$. Thus, we derive a computationally-efficient realization of the EPFES by modeling the
direct-path component of the nonlinear echo path as latent state vector (note that this has been denoted in \cite{saepfes} as SA-\ac{EPFES} to differentiate it from modeling the entire nonlinear echo path) based on~...
\begin{itemize}
 \item the latent length-$3$ vector $\mathbf{a}_n$ modeling the coefficients of the preprocessor and
 \item the latent length-$M_\text{p}$ vector $\mathbf{h}_{\text{p},n}$ modeling the direct-path component of the
\ac{RIR} vector (to circumvent estimation errors of the \ac{NLMS} adaptation in~(\ref{equ:NLMS})).
\end{itemize}
Consequently, the state vector is defined as
\begin{equation}
\mathbf{z}_n=[\mathbf{a}_n^\text{T},\mathbf{h}_{\text{p},n}^\text{T}]^\text{T}
\label{equ:zGeneral}
\end{equation}
and thus of length $M_\text{z}=M_\text{p}+3$ to model the direct-path component of the nonlinear echo path. \\This leads to the observation equation
\begin{equation}
d_{\text{p},n} = \mathbf{g}\left( \mathbf{z}_n \right) + v_n = \mathbf{h}^\text{T}_{\text{p},n} \mathbf{f}\left(\mathbf{x}_{\text{p},n},\mathbf{a}_n \right) + v_n,
\label{equ:ObsAEC}
\end{equation}
precluding a closed-form analytical derivation of the \ac{MMSE} estimate $\mathbf{\hat{z}}_n$~\cite{saepfes}. At this point, we like to emphasize the appropriateness of the observation model in~(\ref{equ:ObsAEC}): The length $M_\text{p}$ of the direct-path component $\mathbf{h}_{\text{p},n}$ is a parameter which can be modified to investigate the system identification performance of the \ac{EPFES} for different sizes $M_\text{z}$ of the search space (according to the definition of the state vector in~(\ref{equ:zGeneral})).\\
As we do not assume to have any priori knowledge about the temporal evolution of the state vector, the process equation is defined as
\begin{equation}
\mathbf{z}_n=\mathbf{z}_{n-1} + \mathbf{w}_n \quad \text{with} \quad \mathbf{w}_n \sim \mathcal{N} \{\mathbf{0},C_{\mathbf{w}} \mathbf{I} \}, \;\; C_{\mathbf{w}}=0.01,
\label{equ:ProcAEC}
\end{equation}
where the state vector $\mathbf{z}_n$ is modeled to be time-invariant up to an additive, zero-mean uncertainty $\mathbf{w}_n$.
%
%
%
\psfrag{s0}[c][c]{(9)}%
\psfrag{s1}[c][c]{$\mathbf{x}_n$}%
\psfrag{s1a}[c][c]{$\mathbf{x}_{\text{p},n}$}%
\psfrag{s1b}[c][c]{$\mathbf{x}_{\text{c},n}$}%
\psfrag{s2}[cc][cc]{$\mathbf{f}(\mathbf{x}_n,\mathbf{\hat{a}}_n)$}%
\psfrag{s3a}[c][c]{$\mathbf{\hat{h}}_{\text{p},n}$ }%
\psfrag{s3b}[c][c]{$\mathbf{\hat{h}}_{\text{c},n}$ }%
\psfrag{s4}[c][c]{$e_n$ }%
\psfrag{s4b}[c][c]{$d_{\text{p},n}$ }%
\psfrag{s5}[c][c]{$d_n$ }%
\psfrag{s5a}[c][b]{$\hat{d}_{\text{p},n}$}%
\psfrag{s5b}[c][b]{$\hat{d}_{\text{c},n}$}%
\psfrag{s6}[c][c]{\small Non-}%
\psfrag{s6b}[c][c]{\small linearities}%
\psfrag{s7}[c][c]{$f(\mathbf{x}_n,\mathbf{a}_n)$}%
\psfrag{s8}[c][c]{$\mathbf{h}_n$}%
\psfrag{s8b}[c][c]{}%
\psfrag{s9}[c][c]{\small Uncertainty}%
\psfrag{s10}[c][c]{}%
\psfrag{s10b}[c][c]{$v_n$}%
\psfrag{s11}[c][c]{SA-EPFES}%
\psfrag{s12}[c][c]{$\mathbf{z}_n=[\mathbf{h}_{\text{p},n}^T , \mathbf{a}_n^T]^T$}%

%
%

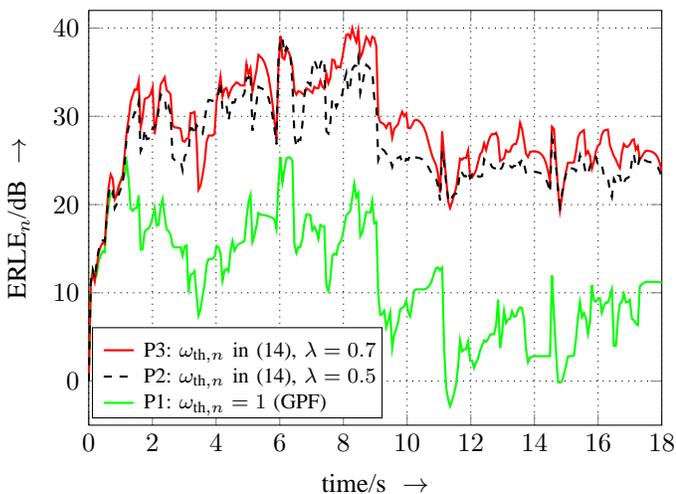
\begin{figure}[!h]
\centering
\begin{tikzpicture}[scale=1]
\def\lx{0.05}
\def\ly{0.1}
\begin{axis}[
      width=9.2cm,height=7.1cm,grid=major,grid style = {dotted,black},
      ylabel style={yshift=-3mm},
      ylabel={$\text{ERLE}_n$/dB $\; \rightarrow$},
      xlabel={time/s $\; \rightarrow$},
      ymin=-5, ymax=42,xmin=0,xmax=18,
      ]
      \addplot[thick,green,solid] table [x index=0, y index=1]{ERLE_GPF41L.dat};
      \addplot[thick,red,solid] table [x index=0, y index=1]{ERLE_SAE41L.dat};
            \addplot[thick,black,dashed] table [x index=0, y index=1]{ERLE_SAEopt41L.dat};
 \end{axis}
 \draw[fill=white] (\lx,\ly) rectangle (3.85+\lx,1.2+\ly);
 \draw[green,thick,solid] (0.15+\lx,0.2+\ly) -- +(0.4,0) node[anchor=mid west,black] {\footnotesize  P1: $\omega_{\text{th},n}=1$ (GPF)};
 \draw[black,thick,dashed] (0.15+\lx,0.55+\ly) -- +(0.4,0) node[anchor=mid west,black] {\footnotesize P2: $\omega_{\text{th},n}$ in (\ref{equ:THmean}), $\lambda=0.5$};
  \draw[thick,red,solid] (0.15+\lx,0.9+\ly) -- +(0.4,0) node[anchor=mid west,black] {\footnotesize P3: $\omega_{\text{th},n}$ in (\ref{equ:THmean}), $\lambda=0.7$};
 \end{tikzpicture}
 \caption{Simulation results for the $\text{ERLE}_n$ defined in~(\ref{equ:MeasuresAEC}) for three different choices of the parameters $\omega_{\text{th},n}$ and $\lambda$, where $\omega_{\text{th},n}=1$ represents the Gaussian particle filter (GPF), $M_\text{z}=44$ and $L=20$ (Configuration C3).} 
  \label{fig:ERLE3} 
\end{figure}

\noindent The experimental setup consists of a synthesized scenario using a measured \ac{RIR} vector $\mathbf{h}$ at a sampling rate of $16$~kHz, additive white Gaussian noise $v_n$ with a long-term SNR of $10$~dB and $18$ seconds of female speech as nonstationary input signal. The nonlinear loudspeaker signal distortion is chosen as
\begin{equation}
\mathbf{f}(\mathbf{x}_n) =   1/4 \cdot \text{tanh} (4 \mathbf{x}_n),
\end{equation}
which has been employed in~\cite{shimauchi2012,saepfes} and leads to a linear-to-nonlinear power ratio of around 9~dB.\\
As evaluation measure, we choose the time-dependent echo return loss enhancement ($\text{ERLE}_n$)
\begin{equation}
\text{ERLE}_n =  10 \log_{10} \left( \mathcal{E} \{ d^2_n \} / \mathcal{E}\{e^2_n\} \right)
\label{equ:MeasuresAEC}
\end{equation}
and stop the filter adaptation after $9$~seconds. This allows to evaluate the online performance during the first half of the signal and should illustrate
how well the instantaneous solution generalizes during the second half, which is an indicator for the actual system identification performance.
For a fair comparison, we fix the parameters of the \ac{NLMS}~adaptation for all experiments~to
\begin{equation}
M=254, \;\;\; \mu=0.5, \;\;\; \epsilon=10^{-4},
\end{equation}
and set $\mathbf{\hat{a}}_n=[0,0,0]^\text{T}$ during the first $0.1$ seconds to facilitate an initial convergence of the \ac{NLMS}~algorithm. Afterwards, we employ the $\text{ERLE}_n$ defined in~(\ref{equ:MeasuresAEC}) to
investigate the impact of the preprocessor estimation (by applying the \ac{EPFES}) for the identification of the Hammerstein system.\\[1mm]
The following evaluation is split into three parts, where each subsection considers a different \ac{EPFES} configuration dependent on the length $M_\text{z}$ of the latent state vector $\mathbf{z}_n$ and the number of $L$ particles:
\begin{itemize}
 \item Configuration C1: $M_\text{z}=14$, $L=100$\\[-3mm]
 \item Configuration C2: $M_\text{z}=44$, $L=100$\\[-3mm]
 \item Configuration C3: $M_\text{z}=44$, $L=20$\\[-2mm]
\end{itemize}
For each of these EPFES configurations C1, C2 and C3, we compare three \ac{EPFES} parametrizations P1, P2 and P3 dependent on the threshold $\omega_{\text{th},n}$ and the smoothing factor $\lambda$:
\begin{itemize}
 \item Parametrization P1: $\omega_{\text{th},n}=1$ (Gaussian particle filter)\\[-3mm]
 \item Parametrization P2: $\omega_{\text{th},n}$ in~(\ref{equ:THmean}), $\lambda=0.5$\\[-3mm]
 \item Parametrization P3: $\omega_{\text{th},n}$ in~(\ref{equ:THmean}), $\lambda=0.7$\\[3mm]
\end{itemize}
\hspace*{3mm}\textit{Configuration~C1~($M_\text{z}=14$, $L=100$)}: The $\text{ERLE}_n$ for the described realizations of the \ac{EPFES} is shown in Fig.~\ref{fig:ERLE}, where we can observe an improved system identification performance with respect to the Gaussian particle filter ($\omega_{\text{th},n}=1$) by including the evolutionary selection process.
The increased performance during the first $9$ seconds highlights the excellent tracking abilities of the \ac{EPFES},
while the improved system identification after $9$ seconds states that the EPFES generalizes the instantaneous solution of the optimization problem very well.
Interestingly, the system identification improves with increasing value of the smoothing factor $\lambda$.
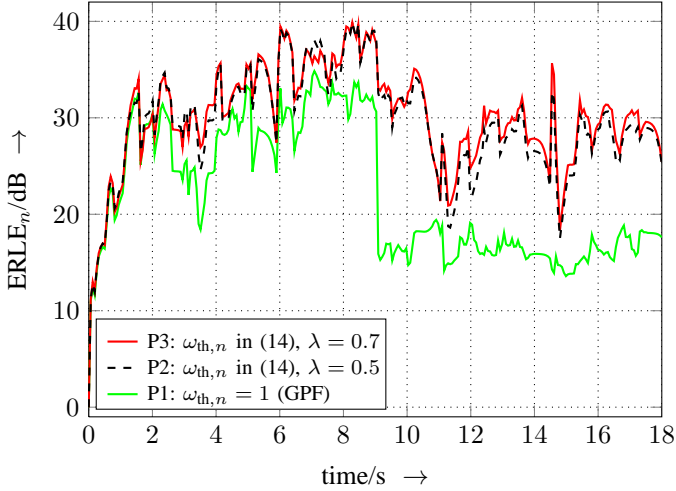
\begin{figure}[!t]
\centering
\begin{tikzpicture}[scale=1]
\def\lx{0.1}
\def\ly{0.1}
\begin{axis}[
      width=9.2cm,height=7.1cm,grid=major,grid style = {dotted,black},
      ylabel style={yshift=-3mm},
      ylabel={$\text{ERLE}_n$/dB $\; \rightarrow$},
      xlabel={time/s $\; \rightarrow$},
      ymin=-1, ymax=42,xmin=0,xmax=18,
      ]
      \addplot[thick,green,solid] table [x index=0, y index=1]{ERLE_GPF.dat};
      \addplot[thick,red,solid] table [x index=0, y index=1]{ERLE_SAE.dat};
            \addplot[thick,black,dashed] table [x index=0, y index=1]{ERLE_SAEopt.dat};
 \end{axis}
 \draw[fill=white] (\lx,\ly) rectangle (3.85+\lx,1.2+\ly);
 \draw[green,thick,solid] (0.15+\lx,0.2+\ly) -- +(0.4,0) node[anchor=mid west,black] {\footnotesize  P1: $\omega_{\text{th},n}=1$ (GPF)};
 \draw[black,thick,dashed] (0.15+\lx,0.55+\ly) -- +(0.4,0) node[anchor=mid west,black] {\footnotesize P2: $\omega_{\text{th},n}$ in~(\ref{equ:THmean}), $\lambda=0.5$};
  \draw[thick,red,solid] (0.15+\lx,0.9+\ly) -- +(0.4,0) node[anchor=mid west,black] {\footnotesize P3: $\omega_{\text{th},n}$ in~(\ref{equ:THmean}), $\lambda=0.7$};
 \end{tikzpicture}
 \caption{Simulation results for the $\text{ERLE}_n$ defined in~(\ref{equ:MeasuresAEC}) for three different choices of the parameters $\omega_{\text{th},n}$ and $\lambda$, where $\omega_{\text{th},n}=1$ represents the Gaussian particle filter (GPF), $M_\text{z}=14$ and $L=100$ (Configuration C1).} 
  \label{fig:ERLE} 
\end{figure}
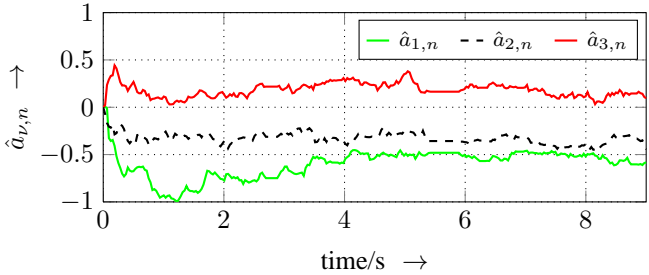
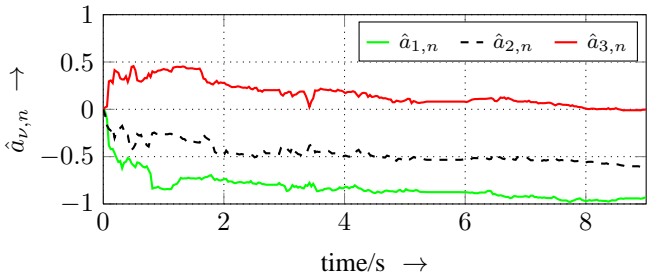
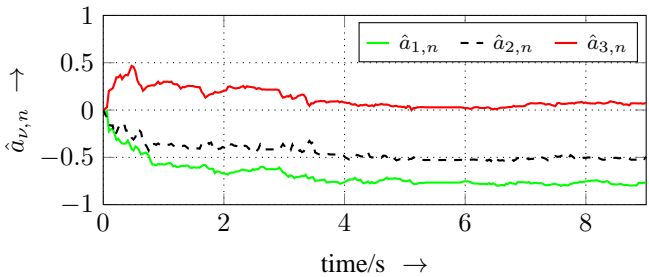
\begin{figure}[!h]
\centering
\subfigure[Simulation results for P1: $\omega_{\text{th},n}=1$ (Gaussian particle filter)]{
\begin{tikzpicture}[scale=1]
\def\lx{3.4}
\def\ly{1.9}
\begin{axis}[
      width=8.8cm,height=4.1cm,grid=major,grid style = {dotted,black},
      ylabel style={yshift=-1mm},
      ylabel={$\hat{a}_{\nu,n}  \; \rightarrow$},
      xlabel={time/s $\; \rightarrow$},
      ymin=-1, ymax=1,xmin=0,xmax=9,
      ]
      \addplot[thick,green,solid] table [x index=0, y index=1]{a1_GPF.dat};
      \addplot[thick,red,solid] table [x index=0, y index=1]{a2_GPF.dat};
            \addplot[thick,black,dashed] table [x index=0, y index=1]{a3_GPF.dat};
 \end{axis}
 \draw[fill=white] (\lx,\ly) rectangle (3.7+\lx,0.52+\ly);
 \draw[green,thick,solid] (0.1+\lx,0.25+\ly) -- +(0.3,0) node[anchor=mid west,black] {\footnotesize $\hat{a}_{1,n}$ };
 \draw[black,thick,dashed] (1.35+\lx,0.25+\ly) -- +(0.3,0) node[anchor=mid west,black] {\footnotesize $\hat{a}_{2,n}$};
  \draw[thick,red,solid] (2.6+\lx,0.25+\ly) -- +(0.3,0) node[anchor=mid west,black] {\footnotesize $\hat{a}_{3,n}$};
 \end{tikzpicture}}\vspace{1mm}
\subfigure[Simulation results for P2: $\omega_{\text{th},n}$ in~(\ref{equ:THmean}), $\lambda=0.5$]{
\begin{tikzpicture}[scale=1]
\def\lx{3.4}
\def\ly{1.9}
\begin{axis}[
      width=8.8cm,height=4.1cm,grid=major,grid style = {dotted,black},
      ylabel style={yshift=-1mm},
      ylabel={$\hat{a}_{\nu,n} \; \rightarrow$},
      xlabel={time/s $\; \rightarrow$},
      ymin=-1, ymax=1,xmin=0,xmax=9,
      ]
      \addplot[thick,green,solid] table [x index=0, y index=1]{a1_SAEopt.dat};
      \addplot[thick,red,solid] table [x index=0, y index=1]{a2_SAEopt.dat};
            \addplot[thick,black,dashed] table [x index=0, y index=1]{a3_SAEopt.dat};
 \end{axis}
 \draw[fill=white] (\lx,\ly) rectangle (3.7+\lx,0.52+\ly);
 \draw[green,thick,solid] (0.1+\lx,0.25+\ly) -- +(0.3,0) node[anchor=mid west,black] {\footnotesize $\hat{a}_{1,n}$ };
 \draw[black,thick,dashed] (1.35+\lx,0.25+\ly) -- +(0.3,0) node[anchor=mid west,black] {\footnotesize $\hat{a}_{2,n}$};
  \draw[thick,red,solid] (2.6+\lx,0.25+\ly) -- +(0.3,0) node[anchor=mid west,black] {\footnotesize $\hat{a}_{3,n}$};
 \end{tikzpicture}}\vspace{1mm}
\subfigure[Simulation results for P3: $\omega_{\text{th},n}$ in~(\ref{equ:THmean}), $\lambda=0.7$]{
\begin{tikzpicture}[scale=1]
\def\lx{3.4}
\def\ly{1.9}
\begin{axis}[
      width=8.8cm,height=4.1cm,grid=major,grid style = {dotted,black},
      ylabel style={yshift=-1mm},
      ylabel={$\hat{a}_{\nu,n} \; \rightarrow$},
      xlabel={time/s $\; \rightarrow$},
      ymin=-1, ymax=1,xmin=0,xmax=9,
      ]
      \addplot[thick,green,solid] table [x index=0, y index=1]{a1_SAE.dat};
      \addplot[thick,red,solid] table [x index=0, y index=1]{a2_SAE.dat};
            \addplot[thick,black,dashed] table [x index=0, y index=1]{a3_SAE.dat};
 \end{axis}
 \draw[fill=white] (\lx,\ly) rectangle (3.7+\lx,0.52+\ly);
 \draw[green,thick,solid] (0.1+\lx,0.25+\ly) -- +(0.3,0) node[anchor=mid west,black] {\footnotesize $\hat{a}_{1,n}$ };
 \draw[black,thick,dashed] (1.35+\lx,0.25+\ly) -- +(0.3,0) node[anchor=mid west,black] {\footnotesize $\hat{a}_{2,n}$};
  \draw[thick,red,solid] (2.6+\lx,0.25+\ly) -- +(0.3,0) node[anchor=mid west,black] {\footnotesize $\hat{a}_{3,n}$};
 \end{tikzpicture}}\vspace{1mm}
 \caption{Estimated coefficients of the preprocessor in~(\ref{equ:y}) for three different choices of the parameters $\omega_{\text{th},n}$ and $\lambda$, where $\omega_{\text{th},n}=1$ represents the Gaussian particle filter, $M_\text{z}=14$ and $L=100$ (Configuration C1).}
  \label{fig:ResultsRPFE}
 \end{figure}
 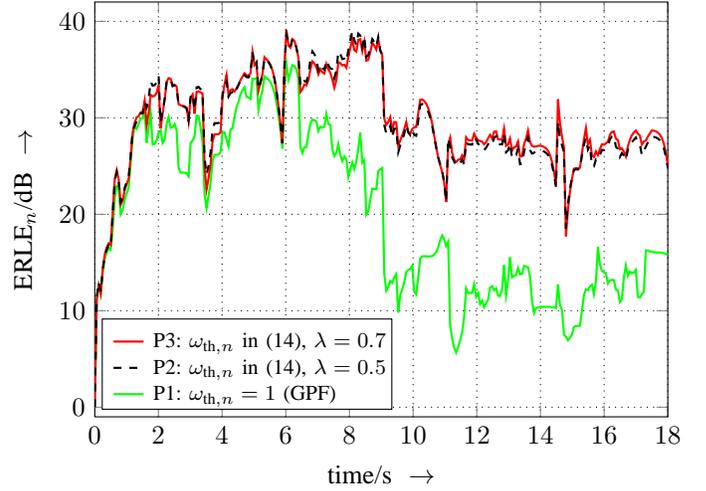
\begin{figure}[!t]
\centering
\begin{tikzpicture}[scale=1]
\def\lx{0.1}
\def\ly{0.1}
\begin{axis}[
      width=9.2cm,height=7.1cm,grid=major,grid style = {dotted,black},
      ylabel style={yshift=-3mm},
      ylabel={$\text{ERLE}_n$/dB $\; \rightarrow$},
      xlabel={time/s $\; \rightarrow$},
      ymin=-1, ymax=42,xmin=0,xmax=18,
      ]
      \addplot[thick,green,solid] table [x index=0, y index=1]{ERLE_GPF41.dat};
      \addplot[thick,red,solid] table [x index=0, y index=1]{ERLE_SAE41.dat};
            \addplot[thick,black,dashed] table [x index=0, y index=1]{ERLE_SAEopt41.dat};
 \end{axis}
 \draw[fill=white] (\lx,\ly) rectangle (3.85+\lx,1.2+\ly);
 \draw[green,thick,solid] (0.15+\lx,0.2+\ly) -- +(0.4,0) node[anchor=mid west,black] {\footnotesize  P1: $\omega_{\text{th},n}=1$ (GPF)};
 \draw[black,thick,dashed] (0.15+\lx,0.55+\ly) -- +(0.4,0) node[anchor=mid west,black] {\footnotesize P2: $\omega_{\text{th},n}$ in (\ref{equ:THmean}), $\lambda=0.5$};
  \draw[thick,red,solid] (0.15+\lx,0.9+\ly) -- +(0.4,0) node[anchor=mid west,black] {\footnotesize P3: $\omega_{\text{th},n}$ in (\ref{equ:THmean}), $\lambda=0.7$};
 \end{tikzpicture}
 \caption{Simulation results for the $\text{ERLE}_n$ defined in~(\ref{equ:MeasuresAEC}) for three different choices of the parameters $\omega_{\text{th},n}$ and $\lambda$, where $\omega_{\text{th},n}=1$ represents the Gaussian particle filter (GPF), $M_\text{z}=44$ and $L=100$ (Configuration C2).} 
  \label{fig:ERLE2} 
\end{figure}
To investigate this behavior in more detail, we consider the estimated parameter values $\hat{a}_{\nu,n}$ of the memoryless preprocessor during the first $9$~seconds
shown in Fig.~\ref{fig:ResultsRPFE}(a)-(c) (for the parametrizations P1, P2, and P3, respectively).
As can be seen in Fig.~\ref{fig:ResultsRPFE}(a), the 
estimated coefficients $\hat{a}_{\nu,n}$ are highly oscillating over time which implies that the Gaussian particle filter solves the instantaneous solution of the optimization problem without identifying the true values of the coefficients. Comparing this to parametrization~P2 shown in Fig.~\ref{fig:ResultsRPFE}(b), we notice the dynamic changes are still present but smaller in amplitude. 
As a consequence of this, the evolutionary selection of elitist particles provides a much more stable result.
This effect is even more pronounced for increasing $\lambda$, which effectively employs more long-time information for judging particles to be within the set of elitist particles (see the results for parametrization P3 shown in Fig.~\ref{fig:ResultsRPFE}(c)). 
The average $\text{ERLE}_n$ of the Hammerstein system using the \ac{EPFES} ($\omega_{\text{th},n}$ in (\ref{equ:THmean}), $\lambda=0.7$) and the Gaussian particle filter ($\omega_{\text{th},n}=1$) for estimating the memoryless preprocessor are 
shown in the first row of Table~\ref{tab:OverEPFES2}.  Note that the average $\text{ERLE}_n$ might be larger in the second half of the experiment ($9$~s - $18$~s in Table~\ref{tab:OverEPFES2}) due to the initial convergence phase.\\[7mm]
\hspace*{3mm}\textit{Configuration~C2~($M_\text{z}=44$, $L=100$):} In this experiment, we evaluate the impact of an increased length of the state vector on the system identification performance of the Hammerstein system with an \ac{EPFES}-estimated memoryless preprocessor.
Note that the simulation results in Fig.~\ref{fig:ERLE2} confirm the statement of Configuration~C1 that the system identification performance improves with increasing value of~$\lambda$. Interestingly, the increased length of the state vector leads to a significant decrease in the $\text{ERLE}_n$ after freezing the filter coefficients at the time instant of $9$~seconds if the threshold is set to $\omega_{\text{th},n}=1$ (Gaussian particle filter). This indicates a worse generalization of the Gaussian particle filter with increasing search space
which can be successfully prevented by introducing the evolutionary selection process of the EPFES: the comparison of Fig.~\ref{fig:ERLE} and Fig.~\ref{fig:ERLE2}
reveals that the \ac{EPFES} is hardly affected by the increased search space. This underlines the extraordinary suitability of the \ac{EPFES} also for optimization problems in higher-dimensional search spaces and is
also clearly stated by a quantitative comparison of the average $\text{ERLE}_n$ values in the second row of Table~\ref{tab:OverEPFES2}. \clearpage
\hspace*{0mm}\textit{Configuration~C3~($M_\text{z}=44$, $L=20$):} Finally, we investigate the identification of the Hammerstein system for estimating the preprocessor coefficients with a reduced number of $L=20$ particles. As shown in Fig.~\ref{fig:ERLE3}, the $\text{ERLE}_n$ resulting from the Gaussian particle filter ($\omega_{\text{th},n}=1$) is significantly reduced with respect to Fig.~\ref{fig:ERLE2}. Interestingly, the \ac{EPFES} with evolutionary selection process has a remarkable system identification performance comparable to its realization with $L=100$ particles in Configuration~C2. Fore more details, consider the average $\text{ERLE}_n$ for online adaptation and frozen filter coefficients summarized in~Table~\ref{tab:OverEPFES2}.
These results imply that the \ac{EPFES} generalizes very well and robustly even for a small number $L$ of particles, which is an important property for an efficient realization in practical applications.\\[5mm]
\hspace*{3mm}\textit{Summary of Task B:} The system identification performance of an Hammerstein system with \ac{EPFES}-estimated memoryless preprocessor preceding an NLMS-adapted linear \ac{FIR} filter was evaluated
by choosing three configurations C1, C2 and C3 with different values for the length of the state vector and the size of the set of particles. In sum, it could be shown in comparison to the Gaussian particle filter that the \ac{EPFES} ...\\[-3mm]
\begin{itemize}
  \item generalizes the instantaneous solution of the optimization problem very well (cf. C1 in Table~\ref{tab:OverEPFES2}),\\[-3mm]
  \item shows a remarkable system identification performance even for high search spaces (cf. C1 to C2 in Table~\ref{tab:OverEPFES2}),\\[-3mm]
  \item requires only a small set of particles for getting a robust estimate of the state vector (cf. C1 to C3 in~Table~\ref{tab:OverEPFES2}).\\[-2mm]
\end{itemize}
These statements are emphasized by the quantitative comparison of the average $\text{ERLE}_n$ values for instantaneous filter adaptation and frozen filter coefficients (after $9$~seconds) shown in Table~\ref{tab:OverEPFES2}.\\
\begin{table}[!t]
\caption{Average $\text{ERLE}_n$ of the Hammerstein system using the \ac{EPFES} (P3) and the Gaussian particle filter (P1) to estimate the memoryless preprocessor, adaptation stopped after $9$~seconds.}
\centering
\vspace{1mm}
\begin{tabular}[c]{|c|c|c|c|c|}
\hline
Parametrization & \multicolumn{2}{c|}{P3: $\omega_{\text{th},n}$ in (\ref{equ:THmean}), $\lambda=0.7$} & \multicolumn{2}{c|}{P1: $\omega_{\text{th},n}=1$ (GPF)}\\
 \hline
Time frame & $0$~s - $9$~s & $9$~s - $18$~s & $0$~s - $9$~s & $9$~s - $18$~s\\
\hline
Config. C1 & 23.7~dB& 26.1~dB& 22.1~dB& 16.1~dB\\
\hline
Config. C2 & 23.6~dB& 26.0~dB& 22.0~dB& 11.6~dB\\
\hline
Config. C3 & 22.8~dB& 24.9~dB& 14.4~dB& 3.5~dB\\
\hline
\end{tabular}
\label{tab:OverEPFES2}
\end{table}
\newpage
\section{Extensions of the EPFES}
\label{cha:outlook}
So far, the \ac{EPFES} has been derived from the well-known state-space model as a generalization of the Gaussian particle filter.
The proposed evolutionary selection process includes a replacement of non-elitist particles of weights smaller than the threshold~$\omega_{\text{th},n}$.
It seems intuitive that employing a more sophisticated selection procedure could further improve the system identification performance of the \ac{EPFES}.
Therefore, we consider the \ac{EPFES} from a different perspective by comparing its properties to basic features of evolutionary algorithms which are optimization techniques inspired by the natural evolution of biological organisms~\cite{Evol_00}: 
\begin{itemize}
 \item \textit{Selection:} Similar to the \ac{EPFES} described in Section~\ref{cha:EPFESproperties}, evolutionary algorithms realize a selection of population members based on their respective fitness scores. 
Extensive research has been made to improve the selection procedure (e.g., \textit{fitness scaling}~\cite{Evol_85} or \textit{ranking selection}~\cite{Evol_105}).
\item \textit{Replacement:} As the size of the population is constant~\cite{Evol_50}, the selection process of evolutionary algorithms is followed by replacing specific members dependent on their current fitness scores.
 However, the choice of an elitist group with the highest fitness scores (and a replacement of the members with the lowest fitness scores) is a special case following the so-called \textit{replacement-of-the-worst strategy}~\cite{Evol_00}. In the theory of evolutionary algorithms, also different replacement procedures could be applied (with the goal to increase the diversity of the population), where most state-of-the-art techniques can be assigned to the so-called \textit{crowding methods}~\cite{Lozano_2008}. 
\item \textit{Mutation and recombination:} In evolutionary algorithms, the steps of mutation and recombination consider the introduction of innovation and the inheritance of genetic information from parents to children, respectively\mbox{\cite{Evol_39,Evol_48,Evol_47}}. 
\end{itemize}
Besides a smarter selection of elitist-particles (see e.g.~\cite{Evol_24,Evol_43}), we expect an adaptive estimation of the smoothing factor $\lambda$ in~(\ref{equ:weightsNew}) (see e.g.~\cite{lucas_exponentially_1990}) as very promising for further improving the system identification performance of the \ac{EPFES}.\\ \\ \\
\section{Conclusion}
\label{cha:concl}
In this article, we introduced the \ac{EPFES} as a promising approach for nonlinear system identification.
Similar to the classical particle filter, the \ac{EPFES} consists of a set of particles and corresponding weights which represent different
realizations of the latent state vector and their likelihood to be the solution of the optimization problem.
To cope with conceptional disadvantages of the classical particle filter, the \ac{EPFES}
includes an evolutionary selection of elitist particles based on long-term fitness measures.
To introduce innovation, the non-elitist particles are replaced by samples drawn from an approximated continuous posterior PDF which solves the problems of degeneracy and sample impoverishment.
With these properties, the \ac{EPFES} is shown to be a generalized form of the widely-used Gaussian particle filter.\\
In this article, we propose two advancements of the EPFES by introducing an adaptive (instead of manually optimized) threshold for the elitist-particle selection and deriving long-term fitness measures from a theoretical perspective (instead of using an intuitive heuristic motivation~\cite{epfes}).\\
The \ac{EPFES} proved to outperform the Gaussian particle filter for two completely different nonlinear state-space models:
First, we considered the \ac{UNGM} with a scalar state variable (quickly changing over time) to investigate the evolutionary selection process of the \ac{EPFES} without using long-term fitness measures. Second, the task of nonlinear \ac{AEC} has been addressed for evaluating the generalization properties of the \ac{EPFES} even for large search spaces.
Both experiments emphasize the remarkable system identification performance of the \ac{EPFES} and show the optimization possibilities as generalized form of the Gaussian particle filter.
In particular, the \ac{EPFES} generalizes the instantaneous solution of the optimization very well. This is the case not only for large spaces but also for a small set of particles which highlight the efficacy in obtaining robust estimates for an unknown system. \\
In future, the system identification performance of the \ac{EPFES} could be further improved by incorporating well-studied selection and replacement strategies from the field of evolutionary algorithms. Reversely, the use of long-term fitness measures (a feature of the \ac{EPFES}) might also be of interest for extending evolutionary algorithms.
\appendix[Derivation of long-term fitness measures]
As discussed in Section~\ref{cha:EPFESproperties}, it might be of interest for the identification of high-dimensional systems with slowly time-varying components to incorporate information about the previous time instants into the particle evaluation as well. For this task, we consider 
a state vector $\mathbf{z}$ which is time-invariant in the interval $n=1,...,N$ and produces the observations $d_{1:n}$.
The starting point for deriving the particle weights is similar to classical particle filtering by drawing $L$ samples $\mathbf{z}^{(l)}$ from the posterior PDF $p(\mathbf{z}|d_{0})$. 
With the goal to derive long-term fitness measures, we assume all samples $\mathbf{z}^{(l)}$ to be elitist particles and thus not to be replaced during the interval $n=1,...,N$.
Under these assumptions, the posterior PDF $p(\mathbf{z}|d_{1:N})$ can be written as
\begin{align}
 p(\mathbf{z}|d_{0:N}) &= \frac{p(d_{1:N}|\mathbf{z},d_{0})  p(\mathbf{z}|d_{0})}{\int p(d_{1:N}|\mathbf{z},d_{0})  p(\mathbf{z}|d_{0}) d\mathbf{z}}\notag \\[2mm]
 &= \frac{p(d_{1:N}|\mathbf{z})  p(\mathbf{z}|d_{0})}{\int p(d_{1:N}|\mathbf{z})  p(\mathbf{z}|d_{0}) d\mathbf{z}}\notag \\[1mm] 
 &= \sum_{l=1}^L \frac{ \prod\limits_{n=1}^N p(d_{n}| \mathbf{z}=\mathbf{z}^{(l)}) }{\sum\limits_{l=1}^L \prod\limits_{n=1}^N p(d_{n}| \mathbf{z}=\mathbf{z}^{(l)})} \delta(\mathbf{z} - \mathbf{z}^{(l)}) \notag \\[1mm]
&= \sum\limits_{l=1}^L \omega^{(l)}_N \delta(\mathbf{z} - \mathbf{z}^{(l)}),\notag
\end{align}
where we defined the weight $\omega^{(l)}_N$ as
\begin{align}
\omega^{(l)}_N = \frac{ \prod\limits_{n=1}^N p(d_{n}| \mathbf{z}=\mathbf{z}^{(l)}) }{\sum\limits_{l=1}^L \prod\limits_{n=1}^N p(d_{n}| \mathbf{z}=\mathbf{z}^{(l)})}.
\label{equ:App1}
\end{align}
Based on the observation equation in (\ref{equ:ObservGen}) with normally-distributed additive observation uncertainty in (\ref{equ:Gaussv}), we insert the Gaussian PDF
\begin{align}
p(d_{n}| \mathbf{z}=\mathbf{z}^{(l)}) = \mathcal{N}\{  \mathbf{g} \{\mathbf{z}^{(l)}\} ,\sigma_{v,n}^2 \} \notag
\end{align}
into the weight calculation in (\ref{equ:App1}) leading to
 \begin{align} \omega_N^{(l)}= \frac{ \text{exp} \left( \frac{1}{N} \sum\limits_{n=1}^N \tilde{\omega}_n^{(l)}\right) }{\sum\limits_{l=1}^L  \text{exp} \left(\frac{1}{N} \sum\limits_{n=1}^N \tilde{\omega}_n^{(l)}\right) }, \quad \tilde{\omega}_n^{(l)}= \frac{(d_{n} - \mathbf{g} \{\mathbf{z}^{(l)}\})^2}{c_{\sigma,n}},
 \label{equ:App2}
 \end{align}\\[-1mm]
where $c_{\sigma,n} = -(2\sigma_{v,n}^2)/N$ is proportional to the variance $\sigma_{v,n}^2$.
As the exact interval length $N$ is generally unknown in practice, we replace the arithmetic sample mean estimator in the exponential functions of (\ref{equ:App2})
by a recursive estimator (often used in speech signal processing~\cite{AdaptiveFilter,haensler_2004}):
%
\begin{align} &\omega_N^{(l)}= 
  \frac{  \text{exp} \left(\hat{\omega}_n^{(l)} \right) }{\sum\limits_{l=1}^L  \text{exp} \left(\hat{\omega}_n^{(l)} \right) }, \quad  \hat{\omega}_n^{(l)}=\lambda \hat{\omega}_{n-1}^{(l)} + (1-\lambda) \tilde{\omega}_n^{(l)}, 
  \label{equ:App3}
  \end{align}
with equal variance for an independent and identically distributed sequence $\tilde{\omega}_n^{(l)}$ ($n=1,...,N$), which means that the time interval $N$ is related to the smoothing factor $\lambda$ via~\cite{hunter_1986}:
\begin{align}
N = (1+\lambda)/(1-\lambda).
\end{align}
Based on this equation, we realize a weight calculation independent of $N$ by estimating $c_{\sigma,n}$ as follows:
\begin{align}
 c_{\sigma,n} = -2\sigma_{v,n}^2 (1-\lambda)/(1+\lambda).
\end{align}
Finally, note that we initialize $\hat{\omega}_{1}^{(l)}=\tilde{\omega}_{1}^{(l)}$ for the recursive update in (\ref{equ:App3}) and that various ways of selecting the smoothing factor $\lambda$ have been considered in the literature, especially in the context of control charts (see e.g. \cite{lucas_exponentially_1990}).

\ifCLASSOPTIONcaptionsoff
  \newpage
\fi

\atColsBreak{\vskip5pt} 
\bibliographystyle{IEEEtran}
\bibliography{literature}

\end{document}